\documentclass[twoside,english]{elsarticle}
\usepackage[latin9]{inputenc}
\usepackage{geometry}
\usepackage{color}
\usepackage{float}
\usepackage{amstext}
\usepackage{amssymb}
\usepackage{graphicx}
\usepackage{subfigure}
\newcommand{\R}{\mathbb{R}}
\makeatletter

\providecommand{\tabularnewline}{\\}
\floatstyle{ruled}
\newfloat{algorithm}{tbp}{loa}
\providecommand{\algorithmname}{Algorithm}
\floatname{algorithm}{\protect\algorithmname}

\usepackage{algorithm}
\usepackage{algpseudocode}
\usepackage{multirow}

\makeatother

\usepackage{babel}
\begin{document}

\title{Locality Preserving Dense Graph Convolutional Networks with Graph Context-Aware Node Representations}

\author[*]{Wenfeng~Liu}

\author[*]{Maoguo~Gong\corref{cor}}

\ead{gong@ieee.org}

\author[*]{Zedong~Tang}

\author[+]{A.~K.~Qin}

\cortext[cor]{Corresponding author}

\address[*]{School of Electronic Engineering, Key Laboratory of Intelligent Perception and Image Understanding of Ministry of Education, Xidian University,
	Xi'an, Shaanxi Province 710071, China}

\address[+]{Department of Computer Science and Software
	Engineering, Swinburne University of Technology, Melbourne, VIC 3122,
	Australia}

\begin{abstract}
Graph convolutional networks (GCNs) have been widely used for representation learning on graph data, which can capture structural patterns on a graph via specifically designed convolution and readout operations. In many graph classification applications, GCN-based approaches have outperformed traditional methods. However, most of the existing GCNs are inefficient to preserve local information of graphs --- a limitation that is especially problematic for graph classification. In this work, we propose a locality-preserving dense GCN with graph context-aware node representations. Specifically, our proposed model incorporates a local node feature reconstruction module to preserve initial node features into node representations, which is realized via a simple but effective encoder-decoder mechanism. To capture local structural patterns in neighbourhoods representing different ranges of locality, dense connectivity is introduced to connect each convolutional layer and its corresponding readout with all previous convolutional layers. To enhance node representativeness, the output of each convolutional layer is concatenated with the output of the previous layer's readout to form a global context-aware node representation. In addition, a self-attention module is introduced to aggregate layer-wise representations to form the final representation. Experiments on benchmark datasets demonstrate the superiority of the proposed model over state-of-the-art methods in terms of classification accuracy.
\end{abstract}

\begin{keyword}
representation learning \sep graph convolutional network \sep graph classification \sep node representation \sep locality preserving \sep dense connection
\end{keyword}
\maketitle

\section{Introduction}
Graph data \cite{tran2018a} which can capture rich information about individual entities and their relations via nodes and edges connecting two nodes has become an essential data type in many research fields including biology (protein-protein interaction networks) \cite{fout2017protein}, chemistry (molecular/compound graphs) \cite{de2018molgan}, cognition intelligence/cognitive sciences (knowledge graphs) \cite{ji2020a}, social sciences (friendship networks) \cite{hamilton2017inductive,kipf2016semi} and many other research areas \cite{battaglia2016interaction,khalil2017learning,yan2018spatial}. Graphs are not only useful as structured information repositories but also play a key role in modern machine learning tasks. Graph classification \cite{scarselli2009the}, as one of the important task based on graph data, aims to classify a given graph into a certain category. For example, in order to distinguish various graph structures of organic molecules, one would ideally infer and aggregate the whole graph topology (which is organized by individual atoms and their direct bonds) as well as node features (e.g., atom attributes), and use the inference and aggregation information to predict the category of the graphs.

\begin{figure}[h]
	\centering
	\includegraphics[scale=0.2]{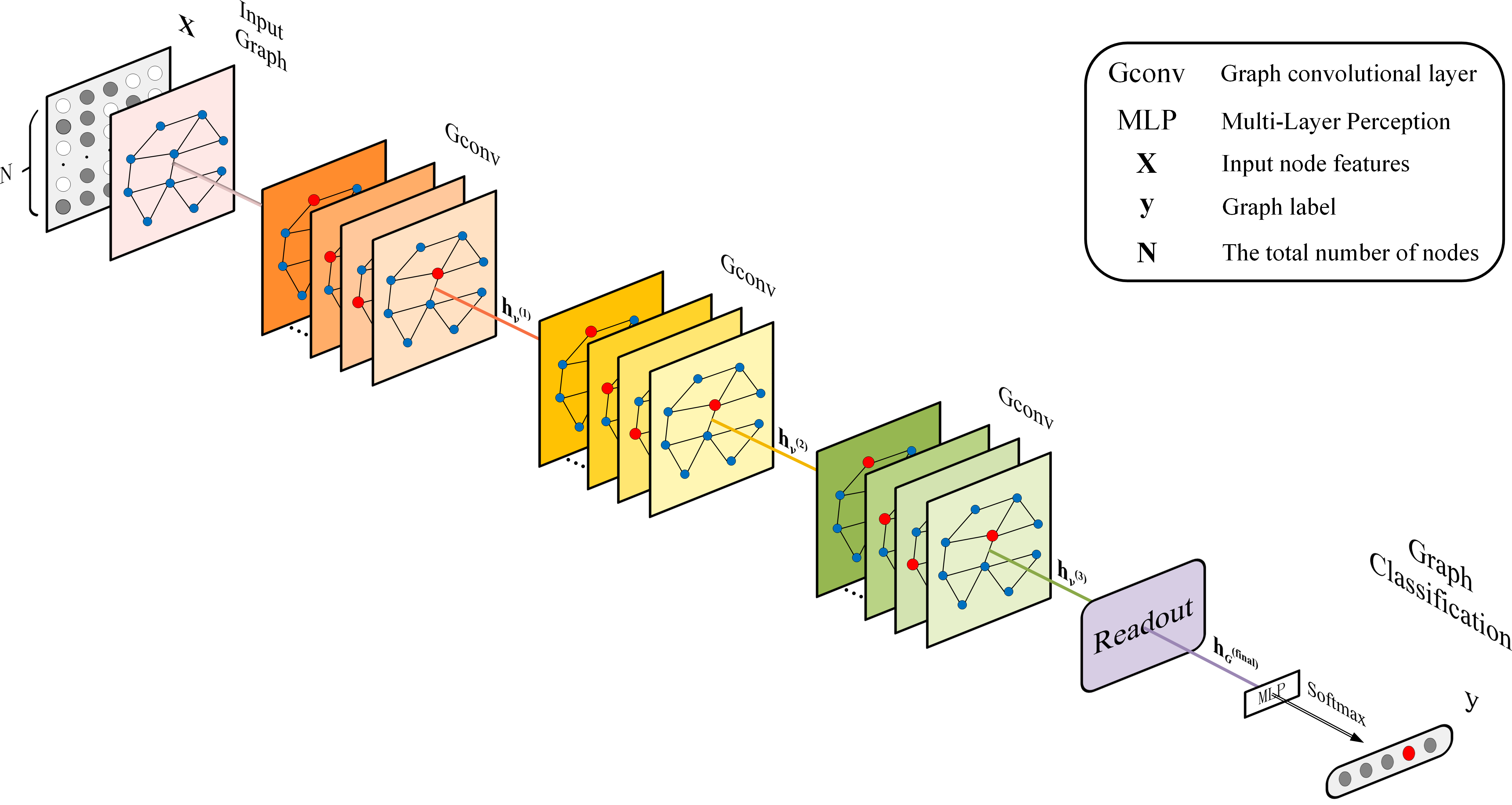}
	\caption{A generic framework of GCNs with 3 convolutional layers and a readout layer for graph classification. A graph convolutional layer encapsulates each node's hidden representation by aggregating feature information from its neighbors. A readout layer collapses node representations of each graph into a graph-level representation.}
	\label{fig:figure1}	
\end{figure}

Recently, a surge of methods have been proposed in the literature aiming to solve the problem of graph classification. Traditionally, A popular technique is to design a kernel function to compute the similarity between pairs of graphs \cite{neumann2016propagation}, followed by a kernel-based classifier like support vector machines (SVM) to perform graph classification. Although effective, kernel-based methods suffer from computation bottleneck and their feature selection is separate from the follow-up classifier \cite{zhao2018substructure}. In order to solve the above challenges, Graph Neural Networks (GNNs), which works in an end-to-end manner, have received increasing research attention \cite{Bacciu2020a,kipf2016semi,xu2018powerful,ying2019gnnexplainer,knyazev2019understanding,li2019deepgcns,Nikolentzos2020k}. Therein, Graph Convolutional Networks (GCNs) are inarguably the dominant category of GNNs to solve the problem of graph classification. Modern GCNs broadly follow the message passing framework \cite{gilmer2017neural} which consists of a message passing phase and a readout phase, in which the message passing phase is to update each node's feature vector by aggregating the feature vectors of its immediate neighbors and the readout phase is then to generate graph-level feature through global pooling modules. GCNs iteratively run $k$-th graph convolution operations using message passing functions to let information travel long distances during the propagation phase. After $k$ iterations, useful features for nodes can be extracted to solve many node-focused analytic tasks (e.g., node classification) \cite{kipf2016semi,cavallari2019embedding}. To tackle graph-focused tasks (e.g., graph classification), the readout module aggregates the information of nodes or local structures to form a graph-level representation \cite{xu2018powerful,zhang2018end}. Figure \ref{fig:figure1} presents a general framework of GCNs for graph classification. Under the message passing framework, many GCN variants with various message functions, vertex update functions, and readout functions have been developed \cite{li2018adaptive,hamilton2017inductive,velivckovic2017graph,kejani2020Graph,Spinelli2020Missing}. 

However, there exists a main limitation of existing GCNs --- their schemes used for graph-level representation learning lack of effective utilization of local feature information. In other words, they overemphasize the capacity in distinguishing different graphs and neglects the local representativeness of a node, leading to over-smoothing (the representation of each node tends to be congruent) when the number of layers becomes large. This is mainly because the local neighborhood aggregation process indiscriminatingly aggregates both positive and negative contributions from the neighbors. As we know, graph-level representation is derived through aggregating local features of nodes, thus locality preserving is the key premise of high graph-level discriminative/representational ability. Existing methods studying on locality preserving for graph-level representation learning can be broadly grouped into three categories: designing differing graph convolution and/or readout operations, hierarchical clustering methods and exploring new model architectures. The first category is to design graph convolution and/or readout operations. For example, in graph isomorphism network (GIN) \cite{xu2018powerful}, Xu et al. discovered that methods under message passing framework are incapable of distinguishing various graph structures based on their extracted graph-level representation, and they proposed an injection neighborhood aggregation scheme and a learnable feature-combine mechanism to preserve local structure and node features of graphs. In \cite{fan2020structured}, Fan et al. proposed a structured self-attention architecture similar to graph attention networks (GATs) \cite{velivckovic2017graph} for graph-level representation learning, where the node-focused attention mechanism aggregates different neighbor node features with learnable weights and the layer-focused as well as the graph-focused self-attention serve as a readout module to aggregate important features from differing nodes and layers to the model's output. In the second category of exploring hierarchical clustering methods, many works prove that graphs also exhibit rich hierarchical structures apart from a dichotomy between node or graph-level structures. A recent advanced work \cite{ying2018hierarchical} proposes DIFFPOOL, a differentiable hierarchical pooling method jointly trained with graph convolutions, to refine local feature information. In summary, the aforementioned two categories of GCN approaches for graph classification are able to overfit most of the training datasets, but their generalization ability is rather limited. In the third category of exploring suitable model architectures, some researchers attempt to remedy the practical difficulty in training GCNs or the over-smoothing problem. In \cite{xu2018representation}, Xu et al. proposed a Jumping Knowledge Network (JK-Net) architecture to connect the last layer of the network with all preceding hidden layers. As a result, the last layer of the model can selectively exploit information from neighborhoods of differing locality, and thus can well capture node-level representation in a fixed number of graph convolution operations. Besides, \cite{pham2017column,rahimi2018semi} introduce residual connections \cite{he2016deep} to skip layers for leveraging local information of different depths, and hence can assist in model training, especially as the depth of the network increases. Such skip connections have been demonstrated to improve the model performance on node-focused tasks, but few works investigated their effectiveness on graph-focused tasks. In GIN, the researchers further proposed a novel architecture similar to JK-Net that concatenates graph-level representations over all layers of the model. Such "readout" architecture that considers all depths of global information can improve generalization of the model.

In this paper, we realize locality preserving in terms of loss function and model architecture. As we know, GCNs utilize the graph topological structure and node features to learn graph-level representation of an entire graph. From the loss perspective, in order to make full use of the node feature information, our model constructs an extra local node-feature reconstruction loss to improve the local representational ability of hidden node representations and enhance the discriminative ability of the final graph-level representation, i.e., we add an auxiliary constraint to preserve locality of the graphs. The local feature reconstruction module is realized by designing a simple but effective encoder-decoder mechanism, in which a stack of multiple graph convolutional layers serves as the encoder and a Multi-Layer Perception (MLP) serves as the decoder. As thus, the input node feature information of the graph is embedded into the hidden representations via the encoder, and then these vector representations are fed into the decoder to reconstruct the initial node features. From the model architecture perspective, we first explore and design a densely connected architecture \cite{huang2017densely} to flexibly leverage information from neighborhoods of differing locality, where we connect each convolutional layer and its readout with all earlier hidden convolutional layers (Note that each convolution module is followed by a readout module in our architecture). Second, we connect each convolutional layer with the previous layer's readout, i.e., the representation outputs of each hidden convolutional layer combine with additional graph-level global context to form global context-aware node representations, aiming at enhancing node representativeness. Additionally, through adding such direct connections from earlier readouts to later convolutions, each readout can receive additional supervision from the local feature reconstruction loss, which is equivalent to impose regularization on the model to reduce the risk of losing local representational ability and hence improve the model's generalization capacity. Lastly, we introduce a self-attention mechanism to aggregate global feature information across different layers to obtain a final graph-level output. Because of the Locality-Preserving Dense GCN architecture, we refer to our model as LPD-GCN.

In summary, our main contributions of this work are three-fold: 
\begin{itemize}
	\item We elaborate an auxiliary local node-feature reconstruction loss by constructing an encoder-decoder mechanism to assist the goal of graph classification. The designed reconstruction module acts as a regularizer, which can help the model to improve the local representational ability of node representations, and thus enhance the global discriminative capacity of the graph-level representation.
	\item We explore a dense connected architecture to flexibly leverage information from neighborhoods of differing locality. Aiming to enhance node representativeness, we connect each convolutional layer with the previous layer's readout. As a result, node representations combine with additional global context to form global context-aware local representations. Moreover, we introduce a self-attention mechanism to adaptively capture all depths of global information. 
	\item Extensive experiments on benchmark datasets indicate that the proposed LPD-GCN can achieve significant improvement of classification accuracies on 4 out of 6 benchmark datasets, which prove its strong representational power, and especially the generalization ability of our model.
\end{itemize} 

The remainder of this paper is organized as follows. We review the relevant research and previous works in Section 2. In Section 3, we discuss the preliminaries of the work. In Section 4, we present our model LPD-GCN in detail. The experimental results and analyses are reported in Section 5. Finally, a conclusion is given in Section 6.

\section{Background and Related Works}
\subsection{Graph Classification and Its Applications}
The goal of graph classification is to distinguish different graph structures based on their learned representations. Once graph representations are derived, the prediction step is equivalent to standard vectorial machine learning, and common loss functions are Cross Entropy/Negative Log-likelihood for classification. A variety of research domains involve the task of graph classification, e.g., biology and chemistry are perhaps the prominent domains where protein structures and chemical compound datasets are often used to benchmark new models on graph classification problems like predicting the functional roles of proteins \cite{fout2017protein}, predicting chemical properties such as toxicity and solubility \cite{gilmer2017neural}. In social science scenario, a challenging graph classification problem is social community classification \cite{yanardag2015a}. The field of computer vision is a promising new application avenue for graph classification. Practical applications including point clouds classification \cite{charles2017pointnet,Lei2020Spherical}, real-time action recognition \cite{yan2018spatial,zhang2020graph}, etc. Existing techniques for graph classification or graph-level representation learning are generally grouped into two main categories: graph kernel-based approaches and graph convolutional network based approaches. In the following, we will briefly review some relevant research on two kinds of approaches.
\subsection{Graph Kernel}
Kernel-based approaches are historically dominant techniques to solve graph classification problem. Informally, a kernel is defined as a generalized form of positive-definite function that measures the similarity scores between pairs of inputs. Specifically, a graph kernel implicitly decomposes a graph into some discrete substructures, such as random walk paths or subtrees, and uses hashing methods to preserve data similarity on feature vectors and measures their similarity score. Graph kernel models following this line of work mainly differ in the way of graph decomposition. For instance, GRAPHLET kernel \cite{shervashidze2009efficient} decomposes a graph into small sub-structures up to a fixed size. Weisfeiler-Lehman (WL) kernel \cite{shervashidze2011weisfeiler} decomposes a graph into subtree patterns. SHORTEST-PATH kernel \cite{borgwardt2005shortest} is based on comparing the paths between graphs. However, the above mentioned graph kernels based on computing the similarity of graphs suffer from the following drawbacks. As the size of substructures increases, the probability that two graphs contain similar substructures will sharply decrease \cite{yanardag2015a}. To resolve this dilemma, a number of methods have been proposed for graph comparison by measuring the differences among the probability distribution of pairwise distances between nodes. For example, \cite{schieber2017quantification} is based on the shortest-path distance, and \cite{verma2017hunt} is based on the diffusion distance. Besides, graph kernels are usually non-adaptive, i.e., they require humans to design proper kernel function to extract features they wanted, which causes the main limitation of kernels. Furthermore, graph kernels suffer from computation bottleneck when the input number in the dataset is too large.

\subsection{Graph Convolutional Network}
In the field of deep neural networks, Scarselli et al. \cite{scarselli2008graph} firstly introduced the concept of GNN, which extends traditional neural networks to deal with irregular non-Euclidean data represented in graph domains. Original GNNs learn node representations through recurrent neural architectures constantly until a stable equilibrium is reached. Lately, graph convolutional networks (GCNs) are inspired by recurrent graph neural networks and inherit the idea of information propagation, which transform traditional convolutional neural networks from Euclidean domain to non-Euclidean graph domain. Advances in this direction are often grouped into spectral-based approaches and spatial-based approaches. Spectral-based approaches define graph convolutions in the Fourier domain by computing the eigendecomposition of the graph Laplacian \cite{bruna2013spectral,kipf2016semi}. Later, \cite{kipf2016semi} builds a bridge between spectral-based and spatial-based approaches, because their convolutional propagation rule is nearly equivalent to the spatial aggregator operation. 

Recently, spatial-based graph convolution, which directly defines operations on node's spatial relations, has been widely adopted in the field of GCNs. The first prominent research on spatial-based convolution is Neural Network for Graphs (NN4G) \cite{micheli2009neural}, which performs graph convolution by directly aggregating information from a node's neighbors. Later, \cite{gilmer2017neural} proposes the Message Passing Neural Network (MPNN) that outlines a unified framework for the graph convolution operation in the spatial domain. It treats graph convolution as a message passing phase (namely, the neighbor aggregation step) in which information can be passed from one node to another along edges directly. The message passing phase iteratively runs $K$-step to let information propagate further. After deriving the hidden representation of each node, MPNN model further runs a readout phase to compute a feature vector for the whole graph. 
A series of neighbor aggregation methods for graph convolutions have been proposed with different ways to aggregate neighbor information. For instance, GraphSAGE \cite{hamilton2017inductive} takes a similar idea as MPNN but performs graph convolution by training an aggregation function that could learn how to aggregate information from a node's local neighborhood. Besides, the authors proposed three different aggregating functions: element-wise mean, long short-term memory (LSTM) and max-pooling. GATs \cite{velivckovic2017graph} introduce the self-attention mechanism into GCNs to learn relative weights of neighboring nodes to the central node, not identical nor pre-defined. Using convolution operations, useful features for nodes can be extracted to solve node-focused analytic tasks. However in \cite{xu2018powerful}, Xu et al. discovered that methods under the MPNN framework(e.g., \cite{kipf2016semi,hamilton2017inductive})) are inefficient to distinguish various graph structures based on their produced graph representations, and they developed a simple architecture GIN, which is as powerful as the Weisfeiler-Lehman graph isomorphism test \cite{shervashidze2011weisfeiler}. Specifically, GIN employs an injection neighborhood aggregation scheme and a learnable feature-combine mechanism to aggregate neighbor information.

Considering graph readout/pooling operations, the most basic operations are simple statistics like taking the sum, mean or max-pooling. However, such primitive reduction operations are inefficient and neglect the impact of node ordering. GIN \cite{xu2018powerful} suggests concatenating graph-level feature information from different layers of the model as a new "readout" architecture to aggregate all depths of global structural information. Moreover, PATCHY-SAN \cite{niepert2016learning} adopts the idea of using a graph labeling procedure to impose an order for neighbor nodes and then resorts to standard 1-D pooling as in CNNs. It can lean from multiple graphs like normal CNNs by preserving node order invariance. SORTINGPOOL \cite{zhang2018end} takes a similar idea to rearrange nodes to a meaningful order, but instead of sorting neighbors for each node, they performs graph pooling by sorting all nodes directly, i.e., a single order for all neighborhoods. Rather than a dichotomy between node or graph-level structures, some researchers suggested making use of rich hierarchical structures in graphs and proposed hierarchical clustering algorithms to generate hierarchical representations of graphs. \cite{ying2018hierarchical} proposed a differentiable graph clustering module (DIFFPOOL) which can be jointly trained with graph convolutions. In each step, DIFFPOOL learns a soft cluster assignment for nodes at each layer of a deep GCN, mapping nodes to several clusters, which then form the coarsened input for the next convolutional layer. After iteratively running $k$ steps, one can learn the feature representation of the whole graph.

\section{Preliminaries}
In this section, we first give the definitions of some key concepts related to our work. Then, we formally introduce the general GCN framework and, along the way, summarize some of the most common GCN variants.
\subsection{Problem Definition}
Throughout this paper, we use $G=\{V,E\}$ to denote the structure of an undirected graph where $V$ represents the set of nodes and $E$ represents the set of edges, $E \subseteq (V \times V)$. Furthermore, we use $X_{v}\in \R^{d_{i}}$ to denote initial node feature for each $v \in V$ assuming each node has $d_i$ features. In this work, we focus on GCN-based graph classification. The aim of GCNs is to learn a continuous representation vector for any graph instance that encodes its node features and its topological structure. Given a set of $M$ labeled graphs $\mathcal{G}= \{G_1,G_2,\cdots,G_M\}$ and their corresponding labels $\mathcal{Y} = \{y_1,y_2,\cdots,y_M\}$, the task of graph classification is to use them as the training data to build a graph classifier $g_{\theta}$ which can assign any new graph input $G$ into a certain class $y_G$.

\subsection{Graph Convolutional Network}
GCNs are one of the most powerful tools for machine learning on graphs, which take into account both the structure information of a graph and the feature information of each node in the graph to learn the node-level and/or graph-level representations that can best assist in the ultimate task to be conducted. Commonly, modern GCNs first aggregate the neighborhood information and then combine the resulting neighborhood representation with the central node's representation from the last iteration. Formally, GCN iteratively updates the representation of a node according to

\begin{equation}\label{aggregate}
h_{v}^{(k)} = {\rm COMBINE}_{k} \Big(h_{v}^{(k-1)}, h_{N(v)}^{(k)} \Big),~ ~ ~ h_{N(v)}^{(k)} = {\rm AGGREGATE}_{k} \Big(\{h_{u}^{(k-1)}: \forall u\in N(v)\} \Big),
\end{equation}%

\noindent where $h_{v}^{(k)} \in \R^{d_{h}}$ denotes the feature representation of node $v$ at the $k$-th iteration with initial input $h_{v}^{(0)}$ = $X_{v}$. ${\rm AGGREGATE}_{k}(\cdot)$ and ${\rm COMBINE}_{k}(\cdot)$ are both learnable information-propagation functions for the $k$-th graph convolutional layer. $N(v) = \{u \in V | (v,u) \in E\}$ denotes the set of adjacent nodes of node $v$. Generally, after $K$ iteration steps, the final node representation $h_{v}^{(K)}$ can be applied for node label prediction, or pass forward to a readout phase performing graph classification. The readout phase aggregates node features to compute a feature vector $h_G$ for the entire graph using some readout functions ${\rm READOUT}(\cdot)$:
\begin{equation}
h_{G} ={\rm READOUT} \Big(h_{v}^{(k)}|v\in G \Big).
\end{equation}
It can be a simple permutation invariant function such as summation or a graph-level pooling function e.g., \cite{ying2018hierarchical,zhang2018end}. The variants of GCNs depend on differing choice of AGGREGATE, COMBINE and/or READOUT functions. For example, in the max-pooling variant of GraphSAGE \cite{hamilton2017inductive}, Hamilton et al. adopted the above idea to define the AGGREGATE and COMBINE steps as
\begin{equation}
h_{v}^{(k)} = \sigma \big(W^{(k)} \cdot [h_{v}^{(k-1)} || h_{N(v)}^{(k)}]\big),~ ~ ~ h_{N(v)}^{(k)} = {\rm MAX} \big(\{\sigma(W_{pool}^{(k)} \cdot h_{u}^{(k-1)} + b) , \forall u\in N(v)\} \big),
\end{equation}
where $W_{pool}^{(k)}$ and $W^{(k)}$ are learnable weight matrices shared by all nodes, $\sigma$ is a non-linear activation function, e.g., a ReLU, and ${\rm MAX}$ denotes an element-wise max-pooling aggregator. $||$ represents the concatenation in the COMBINE step. In GATs \cite{velivckovic2017graph}, Veli{\v{c}}kovi{\'c} et al. incorporated the attention mechanism into the graph aggregate function and computed the representation of each node by applying the multi-head attention mechanism, i.e.,

\begin{equation}
h_{v}^{(k)} = h_{N(v)}^{(k)}, ~ ~ ~ h_{N(v)}^{(k)} = \mathop{||}_{m=1}^M \sigma \big( \sum_{u\in N(v)} \alpha_{vu}^{m} W^{m} h_u \big),
\end{equation}
where $\alpha_{vu}$ denotes the attention coefficient of node $u$ to $v$ and $M$ means the number of independent attention mechanisms. GATs directly take the aggregated neighborhood features to update the central node's representation.
In JK-Net\cite{xu2018representation}, Xu et al. treated the COMBINE step as a form of a "skip connection" between different layers and proposed to use jumping knowledge connections to perform the COMBINE step. Besides, other well-known variants of "skip connections", such as residual connections \cite{he2016deep}, dense connections \cite{huang2017densely} can also be used as the COMBINE step to help information propagation/flow.

In GIN \cite{xu2018powerful}, the element-wise summation is applied as the AGGREGATE step instead, and adjusts the weight of the central node by a learnable parameter $\epsilon^{(k)}$ in the COMBINE step. Formally, the AGGREGATE and COMBINE steps are integrated as follows:
\begin{equation}
h_v^{(k)}={\rm MLP}_{k} \Big(\big((1+\epsilon^{(k)})\cdot h_v^{(k-1)} + \sum_{u\in N(v)} h_u^{(k-1)}\big) \Big),
\end{equation}
where MLP indicates multi-layer perceptrons. For graph-level output, GIN defines a READOUT function to concatenate the graph-level output across all layers in the form of
\begin{equation}
h_{G} = \mathop{||}\limits_{k=0}^{K} \Big( \sum_{v\in G} h_{v}^{(k)} \Big).
\end{equation}

\section{The Proposed Model}
\begin{figure}[h]
	\centering
	\includegraphics[scale=0.16]{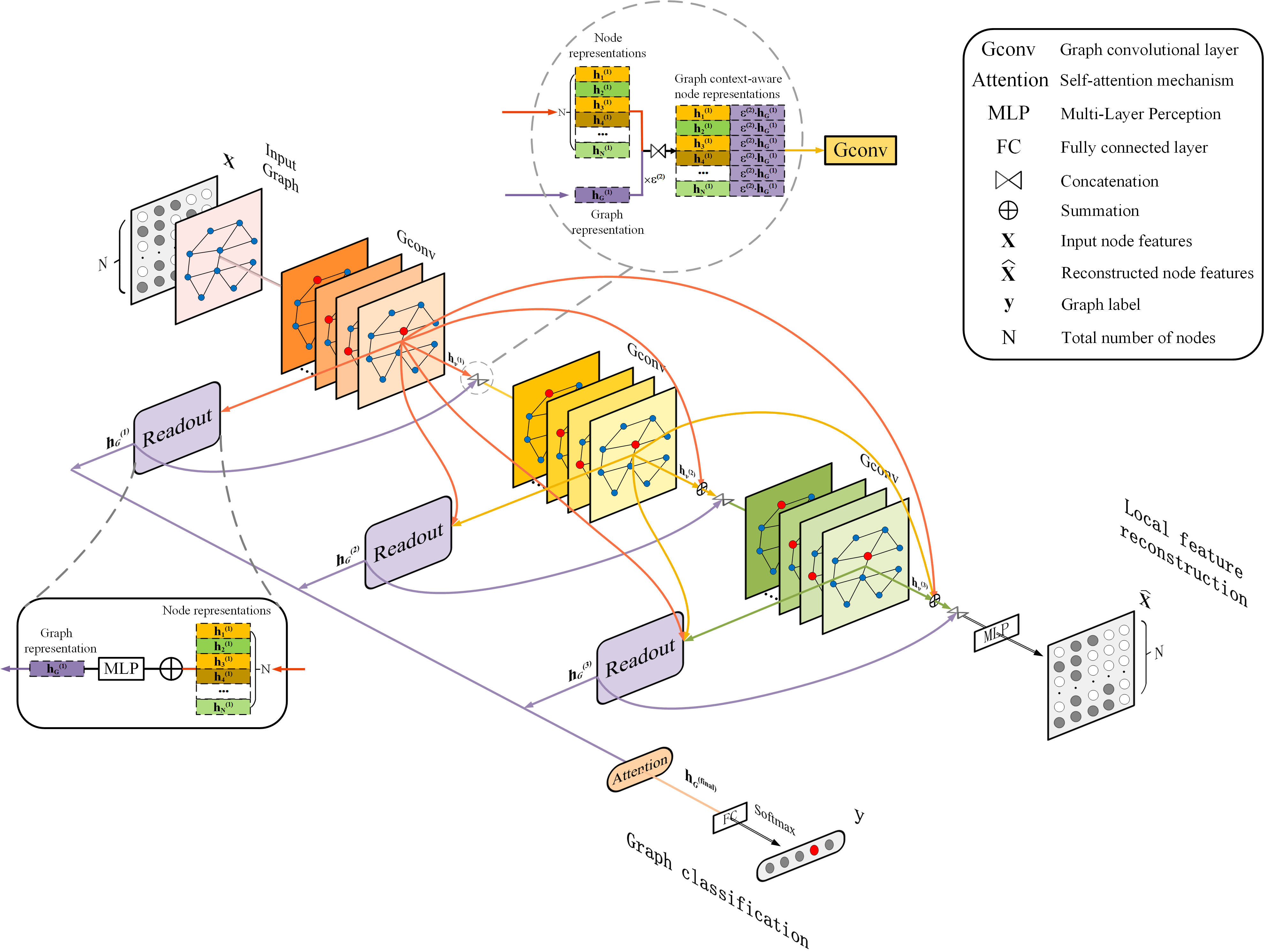}
	\caption{Illustration of a 3-layer LPD-GCN for graph classification. Each layer consists of a graph convolutional layer and a follow-up readout.}
	\label{fig:figure2}	
\end{figure}

This section elaborates our proposed model in the following four parts: (1) \textit{Encoder-decoder based local node-feature reconstruction.} Our model contains a simple encoder-decoder structure to realize local feature reconstruction, in which the encoder is composed of a stack of multiple graph convolutional layers, and the decoder adopts a MLP to reconstruct local node feature. Meanwhile, we construct an auxiliary local feature reconstruction loss to assist the goal of graph classification. As a consequence, the node features can be effectively preserved in the hidden representations over different layers. (2) \textit{Neighborhood aggregation with dense connections.} In our model, we explore a dense connected architecture to flexibly leverage information from neighborhoods of differing locality. That is, we connect each convolutional layer and its corresponding readout with all earlier hidden convolutional layers. (3)\textit{Global context-aware local representations.} We connect each convolutional layer with the previous layer's readout. In this way, each node representation combines with additional global context to form a global context-aware local representation, which can enhance the node representativeness. Besides, each readout can receive additional supervisions from the local feature reconstruction loss, which reduces the risk of losing local representational ability and improve the generalization capacity.
(4)\textit{Attention based layer-wise aggregation.} To effectively extract all depths of global feature information, we further introduce a self-attention mechanism to aggregate layer-wise graph-level feature information of the model. Figure \ref{fig:figure2} illustrates the above main ideas schematically.

\subsection{Encoder-Decoder Based Local Feature Reconstruction}
As we described above, node representations capturing local structure and feature information of graphs are fed to global readouts getting refined and global for graph-level representation learning. However, the graph-level representational ability of traditional GCNs is limited to excessive refinement and globalization and neglect of locality preserving, which causes the over-smoothing problem. To alleviate this problem, we design an encoder-decoder mechanism in our model to reconstruct the feature information of each node in the graph, where a stack of multiple graph convolutional layers is treated as the encoder to encode node feature information into latent vector space, and the decoder employs a simple two-layer MLP to reconstruct initial node features from the encoded vector information. The encoder-decoder mechanism allows the reassessment of initial node features. Afterwards, we impose a local feature reconstruction loss after the decoder, to assist the goal of graph classification. The specific loss function will be detailed in \textit{section 4.5}. 

From the model illustration shown in Figure \ref{fig:figure2}, The encoder that consists of $K$ graph convolutional layers aggregates local neighboring information of different structural levels. After $K$ iterations of neighborhood aggregation, the output of the encoder extracts the local structural information within $K$-hop neighborhood, as well as node feature information. The encoder encodes node feature $X_v \in \R^{d_{i}}$ for each node $v \in V$ into intermediate representations, i.e., $ h_{v}^{(1)}, \cdots, h_{v}^{(k)}, \cdots, h_{v}^{(K)} = {\rm Enc}(X_{v})$, where $h_{v}^{(k)} \in \R^{d_{h}}$. While a decoder aims to map the feature-encoded node representation to a reconstructed feature vector $\widehat{X} \in \R^{d_{i}}$. From Figure \ref{fig:figure2}, given the graph-level representation $h_{G}^{(K)} \in \R^{d_{o}}$ of the readout from the last layer, $\widehat{X}$ can be computed as
\begin{equation}\label{Dec}
\widehat{X} = {\rm Dec} \Big( \big[\sum_{i=1}^{K} h_{v}^{(i)} ~ || ~ h_{G}^{(K)} \big] \Big).
\end{equation}
In our model, we replace the decoder in Equation (\ref{Dec}) with a two-layer MLP. It is worth noting that, our encoder-decoder mechanism is different from the graph autoencoder model \cite{kipf2016variational}. Our decoder tries to reconstruct the initial node features, while their decoder is a task-specific classifier to recover the graph adjacency structure.

\subsection{Neighborhood Aggregation with Dense Connections}

Moreover, to flexibly leverage information from neighborhoods of differing locality, we add direct connections from every hidden convolutional layer to all later convolutional layers and readout layers. The resulting architecture is approximately a graph correspondent of DenseNets, which were proposed for Computer Vision problems \cite{huang2017densely}. Figure \ref{fig:figure2} illustrates this dense connectivity pattern. This architecture allows for selectively combining different aggregations at different layers and further improves the information flow between layers. In DenseNets \cite{huang2017densely}, the layer-wise concatenation aggregation is applied. In our model, we take the layer-wise summation aggregation mechanism instead. Formally, assuming a non-linear injective funtion $\phi(\cdot)$ and an aggregating funtion $f(\cdot)$, the $k$-th convolutional layer aggregates and updates node features iteratively with $h_{v}^{(k)} = \phi(\sum_{i=1}^{k-1} h_{v}^{(i)}, f(\{ \sum_{i=1}^{k-1} h_{u}^{(i)} : u \in N(v) \}))$. In practice, we use a MLP to model $\phi(\cdot)$ and a sum-aggregation function to model $f(\cdot)$. Consequently, our model updates node representations of the $k$-th layer as
\begin{equation}\label{NR1}
h_{v}^{(k)} = {\rm MLP_{k}} \Bigg( \sum_{i=1}^{k-1} \sum_{u\in N(v)} h_{u}^{(i)} \Bigg),
\end{equation}
with $h_{v}^{(0)}=X_{v}$, and especially, $h_{v}^{(1)}={\rm MLP_{1}} \Bigg(\sum_{u \in N(v)} h_{u}^{(0)}\Bigg)$. Among that, $\sum_{u\in N(v)} ( h_{u}^{(i)})$ denotes a neighborhood sum-aggregation and $\sum_{i=1}^{k-1} (\cdot)$ can be viewed as a layer-wise COMBINE step.

\subsection{Global Context-Aware Local Representations}
After introducing the auxiliary local feature reconstruction loss, each convolutional layer can receive additional supervision for locality preserving. However, such supervision information cannot backpropagate to train those global readouts. In our architecture, each convolutional layer is followed by a global readout to collapse node embeddings of an entire graph into a graph-level representation. How to better utilize the supervision information from the local node-feature reconstruction? To tackle this problem, our architecture adds direct connections from each readout to the next convolutional layer and aligns node-level representations with global graph-level representation using concatenation. That is, we use a vertex-wise concatenation that concatenates each node representation with the graph-level representation into a single tensor. Moreover, we introduce a learnable parameter $\epsilon^{(k)} (> 0)$ to adaptively seek a trade-off between the local node-level representation and the global graph-level representation. More formally, given the node representations $\{h_{v}^{(1)}, \cdots, h_{v}^{(k-1)}\}$ and the graph representation $h_{G}^{(k-1)}$, our model now updates node representations of the $k$-th layer as
\begin{equation}
h_{v}^{(k)} = {\rm MLP_{k}} \Bigg( \Big[ a_{v}^{(k-1)}~ \vert\vert ~\epsilon^{(k)} \cdot h_{G}^{(k-1)} \Big]\Bigg),
\end{equation}
where $\vert\vert$ refers to concatenation, $a_{v}^{(k-1)} = \sum_{i=1}^{k-1} \sum_{u\in N(v)} h_{u}^{(i)}$ corresponds to aggregating representations of its neighbors over all its earlier layers according to Equation (\ref{NR1}). Through designing such architecture, besides gradient information from the loss of the main graph-level task, other gradient information from the local feature reconstruction loss can backpropagate to update the parameters of readouts, which can reduce the risk of losing local representational ability and improve the generalization capacity of the model. Meanwhile, node representations combine with additional global context to form global context-aware node representations, which can also enhance node representativeness.

\subsection{Attention Based Global Layer-wise Aggregation}
Traditionally, node embeddings learned by multiple graph convolutional layers are fed to a global readout to yield the graph-level representation, and the readout function generates the graph-level feature through pooling or summation. However, as the depths of the network increases, node embeddings may appear over-smoothing which leads to the poor generalization performance of graph-level output. To effectively extract and utilize all depths of global information, our architecture further employs a self-attention mechanism to aggregate the layer-wise graph-level features of the readouts similar to GIN \cite{xu2018powerful}. The intuition behind introducing a layer-focused self-attention mechanism is that during generating the task-specific graph-level output, the attention weight assigned to each layer can be adaptive to the specific task. In practice, each readout first takes the sum of node embeddings and then feed to a MLP to perform scale transformation. It is worth noting that an extra MLP after summation is necessary because in our architecture each readout is also connected with all preceding convolutional layers. Firstly, each readout updates graph-level representation as
\begin{equation}
h_{G}^{(k)} = {\rm MLP}_{k} \Big( \sum_{v\in G} \sum_{i=1}^{k} h_{v}^{(i)} \Big).
\end{equation}

Afterwards, given the layer-wise graph-level representations of readouts, i.e., $\{h_{G}^{(1)}, \cdots,h_{G}^{(k)}, \cdots, h_{G}^{(K)}\}$, $h_{G}^{(k)} \in \R^{d_{o}}$, the final graph-level output can be computed by using a layer-focused attention mechanism. A self-attention mechanism first computes an attention score $\alpha_{k}$ for each layer $k$ ($\sum_{k=1}^{K} \alpha_{k} = 1$), which represents the importance of the global feature learned on the $k$-th layer. Finally, the final aggregated representation for the graph is a weighted average of the layer features. The detailed formula is
\begin{equation}
h_{G}^{(final)} = \sigma(\sum_{k=0}^{K} \alpha_{k} (W_{1} \cdot h_{G}^{(k)})), ~ ~ ~\alpha_{k} =  {\rm softmax} \Big( W_{2} \cdot \sigma \big(W_{1} \cdot h_{G}^{(k)} \big) \Big),
\end{equation}
where $W_{1},W_{2} \in \R^{d_{o} \times d_{o}}$ are the learnable weight matrices, and a \textit{softmax} function is used for normalization.

\subsection{Loss Functions}

In the training stage, our model receives gradient signals from both the main graph classification task and auxiliary local feature reconstruction regularization. Formally, we train the proposed LPD-GCN by minimizing the total loss defined in Equation (\ref{{TotalLoss}}) weighted by the graph classification loss and the local feature reconstruction loss.
\begin{equation}\label{{TotalLoss}}
\mathcal{L} = \mathcal{L}_{GC} + \lambda \mathcal{L}_{LFR}.
\end{equation}
where $\mathcal{L}_{GC}$ denotes the graph classification loss, and $\mathcal{L}_{LFR}$ denotes the local feature reconstruction loss, the trade-off parameter $\lambda$ is introduced to adaptively seek a balance between the two loss terms. In the following description, we will detail the two objectives, respectively.

Given a batch of graphs $\{G_{1}, \cdots, G_{B}\}$ for training, we first consider the graph classification loss. The output $h_{G}$ at the attention layer is regarded as the final graph representation, followed by a fully connected layer and a softmax non-linear layer that predicts the probability distribution over classes. Consequently, we minimize the cross-entropy of the predicted and true distributions as 
\begin{equation}
\mathcal{L}_{GC} = - \sum_{i=1}^{B} \sum_{j=1}^{C} y_{i,j} \log \hat{y}_{i,j},
\end{equation}
where $y_{i,j}$ is the ground-truth label; $\hat{y}_{i,j}$ is prediction probability; $B$ denotes the batch size for training and $C$ is the number of graph classes. 

As for the local feature reconstruction loss, if the input and reconstructed features are one-hot encoded vectors for node labels, we also stack an extra \textit{softmax} non-linear layer and minimize the cross-entropy loss as
\begin{equation}
\mathcal{L}_{LFR} = - \sum_{i=1}^{B} \sum_{v\in G_i} \sum_{t=1}^{T} X_{v,t} \log \hat{X}_{v,t},
\end{equation}
where $T$ is the number of node classes. If the input and reconstructed node features are continuous-encoded vectors, we shall employ the commonly used \textit{Root Mean Squared Error} (RMSE) metric to evaluate the reconstructed loss between $X_{v}$ and $\hat{X}_{v}$
\begin{equation}
\mathcal{L}_{LFR} = \sqrt{\frac{1}{N_B} \sum_{i=1}^{B} \sum_{v\in G_{i}} (X_{v} - \hat{X}_{v})^2},
\end{equation}
where $N_{B}$ is the total number of nodes in a batch of graphs. In this work, we mainly focus on reconstructing one-hot encoded node label information. We leave the question of continuous-encoded feature reconstruction as our future work.

\section{Experiments}

In this section, we empirically evaluate our model, LPD-GCN, against a number of state-of-the-art baselines on graph classification benchmark datasets, to answer the following questions:
\begin{itemize}
	\item[\textbf{Q1}] How does the proposed LPD-GCN compare to other state-of-the-art GCN variants and kernel-based approaches for graph classification task, with both representational power and generalization capacity?
	\item[\textbf{Q2}] Does the auxiliary node-feature reconstruction objective facilitate the graph classification task?
	\item[\textbf{Q3}] Does the dense connections and global context-aware architecture help improve the performance of the proposed model?
	\item[\textbf{Q4}] How does the selection of hyper-parameters like trade-off parameter $\lambda$ and dropout ratio influence the performance of the proposed model?
\end{itemize}

\noindent\textbf{Datasets.} We use 6 bioinformatics datasets: MUTAG, ENZYMES, PTC, PROTEINS, NCI1, D\&D,\footnote{Download from https://ls11-www.cs.tu-dortmund.de/staff/morris/graphkerneldatasets.} which are balanced graph datasets chosen from benchmarks commonly tested in graph classification. A concise summary are listed in Table \ref{tab:dataset}. Most of the social network datasets lacking of node features are unsuitable for our model. We evaluate the performance via performing 10-fold cross-validation, and report the mean/std-dev of test accuracies.
\begin{itemize}
	\item MUTAG \cite{debnath1991structure} is a dataset of 188 mutagenic aromatic and heteroaromatic nitro compounds dataset, and the graphs are labeled according to whether or not the mutagenic effects on bacteria exists. The MUTAG dataset has 7 discrete node labels.
	\item ENZYMES and PROTEINS \cite{borgwardt2005protein} consist of graph representations of proteins. Nodes represent
	secondary structure elements (SSE), and edges between two nodes represent that they are neighbors in the amino acid sequence or one of three nearest neighbors in space. The discrete attributes are SSE types. The continuous attributes are the 3D length of the SSE. Graph classes indicate which EC top-level class they belong to. The datasets have 3 discrete labels, representing helix, sheet or turn.
	\item PTC \cite{helma2001predictive} is a collection of 344 chemical molecules which report the carcinogenicity for male and female rats. In each graph, nodes represent atoms and edges indicate chemical bonds between nodes. It has 19 discrete labels.
	\item NCI1 \cite{shervashidze2011weisfeiler} is also a chemical compound dataset. It contains 4,110 chemical compounds screened for activity against non-small cell lung cancer and ovarian cancer cell lines with 37 discrete labels.
	\item D\&D \cite{dobson2003distinguishing} consists of graph representations of 1,178 proteins. In each graph, nodes represent amino acids, and there is an edge if they are less than 6 Angstroms apart. Graph classes indicate whether or not they are enzymes. The D\&D dataset has totally 81 discrete labels.
\end{itemize}

\renewcommand{\arraystretch}{1.2}
\begin{table*}[hbt]
	\centering
	\begin{tabular}{ccccccc}
		\hline
		Dataset  & Description & \# Graphs & \# Classes & \# Avg nodes & \# Avg edges & \# Node Labels \\
		\hline
		MUTAG    & Chemical Compounds  & 188  & 2 & 17.93 & 19.79 & 7    \\
		ENZYMES   & Protein Structures    & 600  & 6 & 32.63 &64.14 & 3     \\
		PTC    & Chemical Compounds  & 344  & 2 & 14.29 &14.69 &19     \\
		PROTEINS  & Protein Structures  & 1,113  & 2 & 39.06 & 72.82 & 3     \\
		NCI1  & Chemical Compounds  & 4,110  & 2 & 29.87 & 32.30 & 37     \\
		D\&D  & Protein Structures  & 1,178 & 2 & 284.32 &715.66 & 81 \\
		\hline
	\end{tabular}
	\caption{Statistics of the benchmark graph datasets}
	\label{tab:dataset}
\end{table*}

\noindent\textbf{Models and configurations.} For ablation study, under the LPD-GCN model, we consider the following three variants: (1) LPD-GCN(NoLFR): The LPD-GCN with \underline{No} \underline{L}ocal \underline{F}eature \underline{R}econstruction module, (2) LPD-GCN(NoDC): The proposed LPD-GCN with \underline{No} \underline{D}ense \underline{C}onnections module, and (3) LPD-GCN(NoGCA): The proposed LPD-GCN with \underline{No} \underline{G}raph \underline{C}ontext \underline{A}ware module. The graph classification results of the three variants are also concluded. 

For configurations, we apply 5 graph convolutional layers (including the input layer), and all MLPs have 2 layers. For the first hidden layer, The number of hidden units is set to 64. For the following hidden layers, due to the graph context-aware module involves feature concatenation, the number of hidden units is extended to $2\cdot 64$. Each hidden layer employs a batch normalization (BN) layer \cite{ioffe2015batch} after MLP. For training, we use the Adam optimizer \cite{kingma2015adam} with initial learning rate 0.01 and decay the learning rate by 0.5 every 20 epochs. The exact hyper-parameters are fixed as follows after fine tuning: Minibatch of size $\in\{32,64\}$, and 0.5 dropout ratio, the trade-off parameter $\lambda$ is set to 0.2 in the comparison experiments. Furthermore, we will study on the effects of different choices of trade-off parameter $\lambda$ and dropout ratio for LPD-GCN.

\noindent\textbf{Baselines.} For better comparisons, we consider two categories of state-of-the-art baselines, including GCN-based and kernel-based approaches. 
\begin{itemize}
	\item  \textit{GCN-based approaches} (Semi-GCN\cite{kipf2016semi}, GAT\cite{velivckovic2017graph}, GRAPHSAGE\cite{hamilton2017inductive}, PATCHY-SAN\cite{niepert2016learning}, SORTPOOL\cite{zhang2018end}, DIFFPOOL\cite{ying2018hierarchical}, GIN\cite{xu2018powerful}), which are state-of-the-art GCN variants. For Semi-GCN, GRAPHSAGE, GAT, because they were not applied to graph classification in the original work, we opted for a sum-concat global aggregation function followed \cite{xu2018powerful} to classify graph instances. For PATCHY-SAN, SORTPOOL and DIFFPOOL, they apply a GCN architecture and then perform a graph-level pooling function on node vectors to obtain a representation of an entire graph.
	\item  \textit{Kernel-based approach} (GRAPHLET \cite{shervashidze2009efficient}, SHORTEST-PATH \cite{borgwardt2005shortest}, WL subtree \cite{shervashidze2011weisfeiler}), which are employed as kernel baselines. For GRAPHLET, we set the size of the graphlets to be 7. For WL subtree and SHORTEST-PATH we set the number of iterations $\in \{1,2,\cdots,6\}$. Moreover, we used the \textit{C}-SVM implementation of LIBSVM \cite{chang2011libsvm} to compute the classification accuracies over 10-fold cross validation.
\end{itemize}
Moreover, for baseline methods have reported the published accuracies, we report the accuracies following the original papers. 

\subsection{Results for Graph Classification}
We firstly evaluate the representational power of the proposed LPD-GCN. Then, we provide our theoretical analysis of the representational power of methods by comparing their training accuracies. 

\begin{figure}[htb]
	\centering
	\subfigure[MUTAG]{
		\label{figure3:subfig:a}
		\includegraphics[scale=0.235]{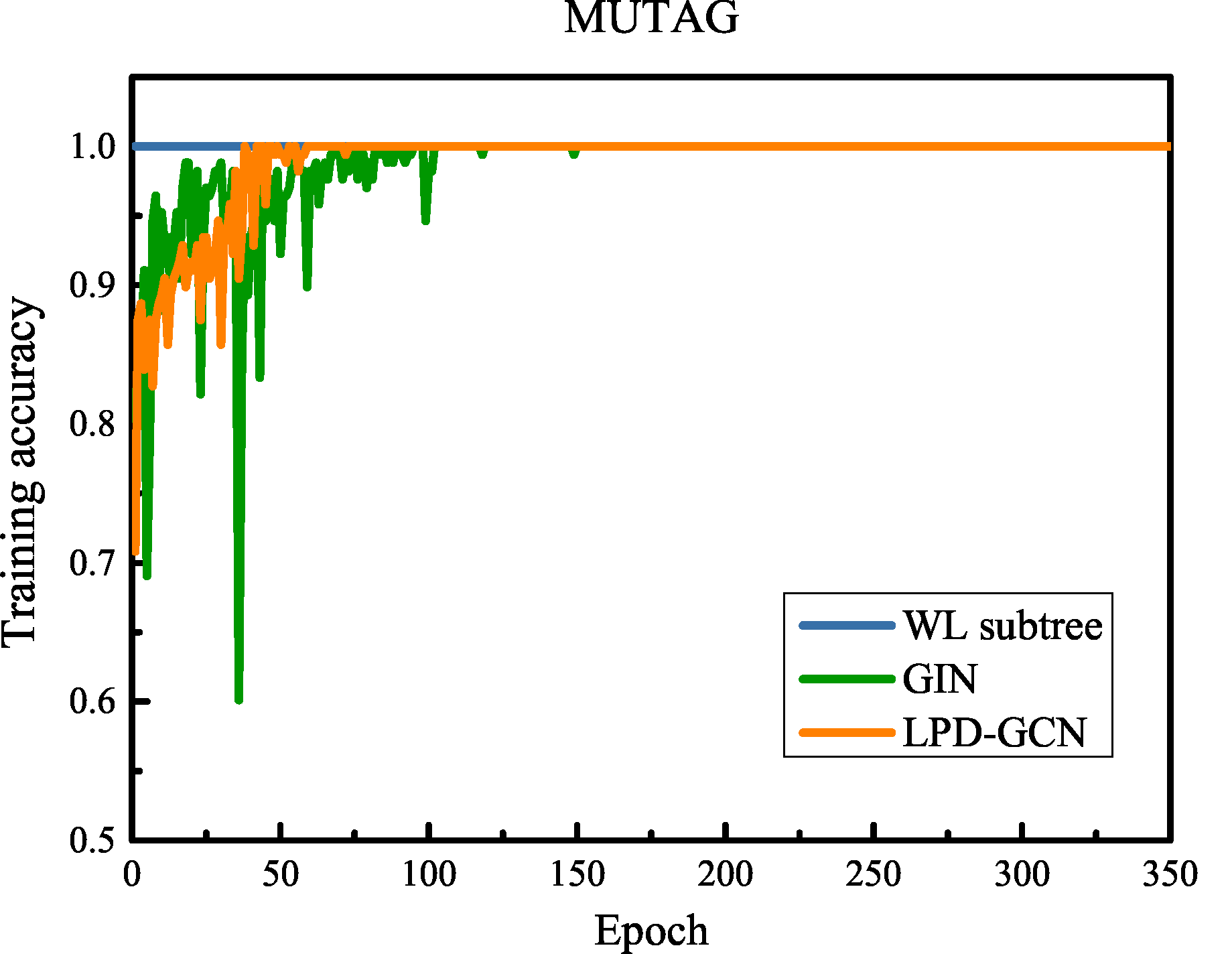}}
	\subfigure[ENZYMES]{
		\label{figure3:subfig:b}
		\includegraphics[scale=0.235]{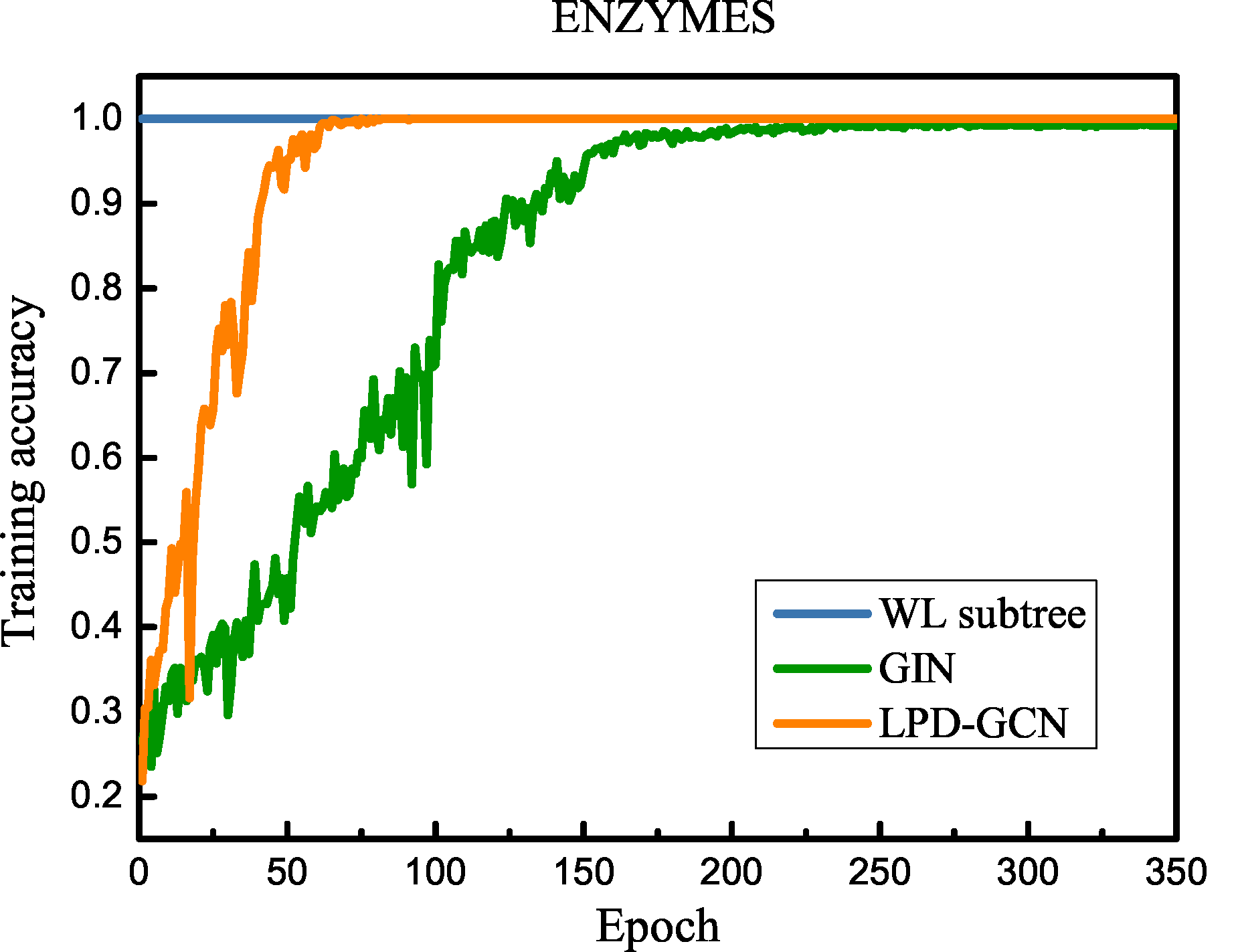}}
	\subfigure[PTC]{
		\label{figure3:subfig:c}
		\includegraphics[scale=0.235]{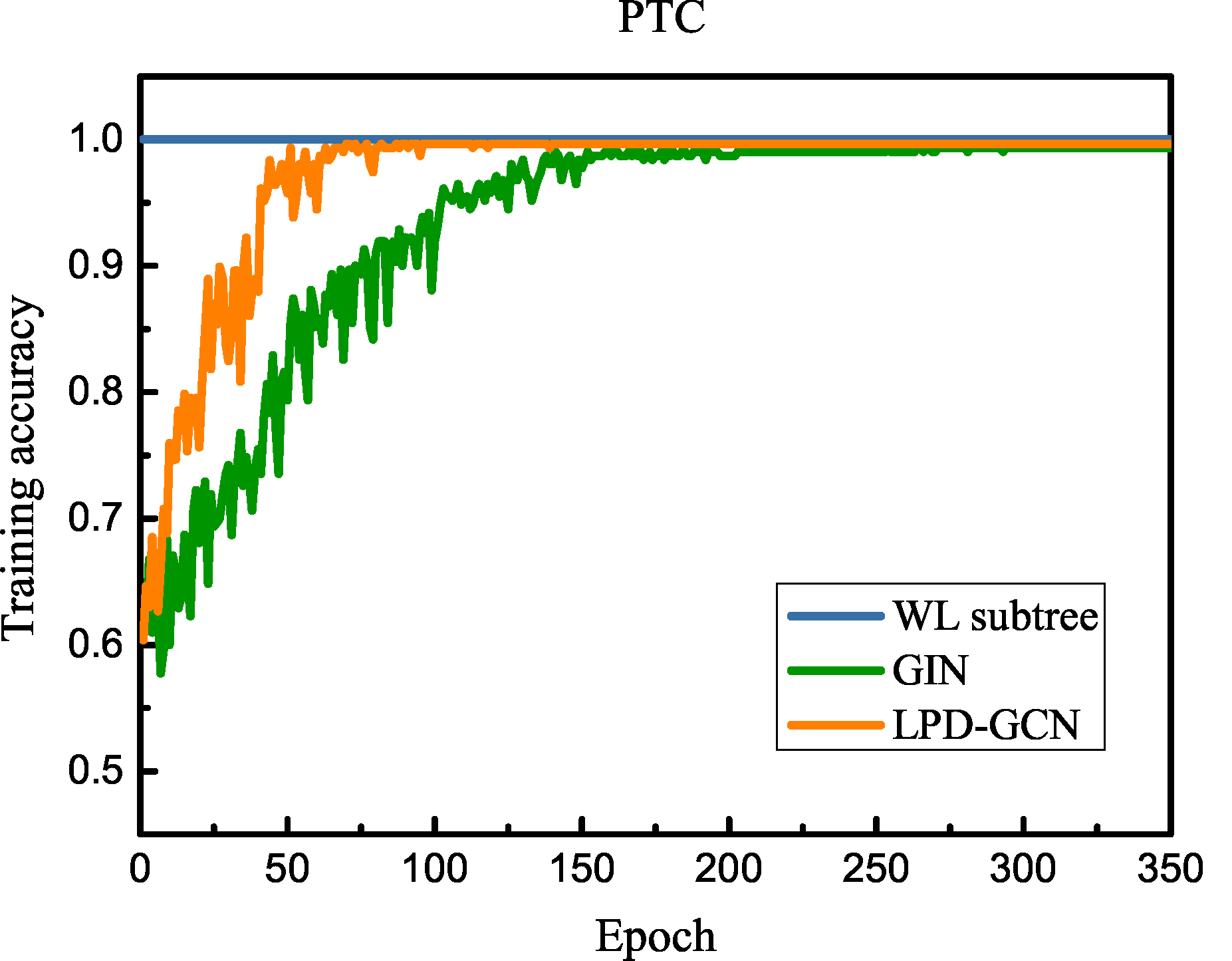}}
	\subfigure[PROTEINS]{
		\label{figure3:subfig:d}
		\includegraphics[scale=0.235]{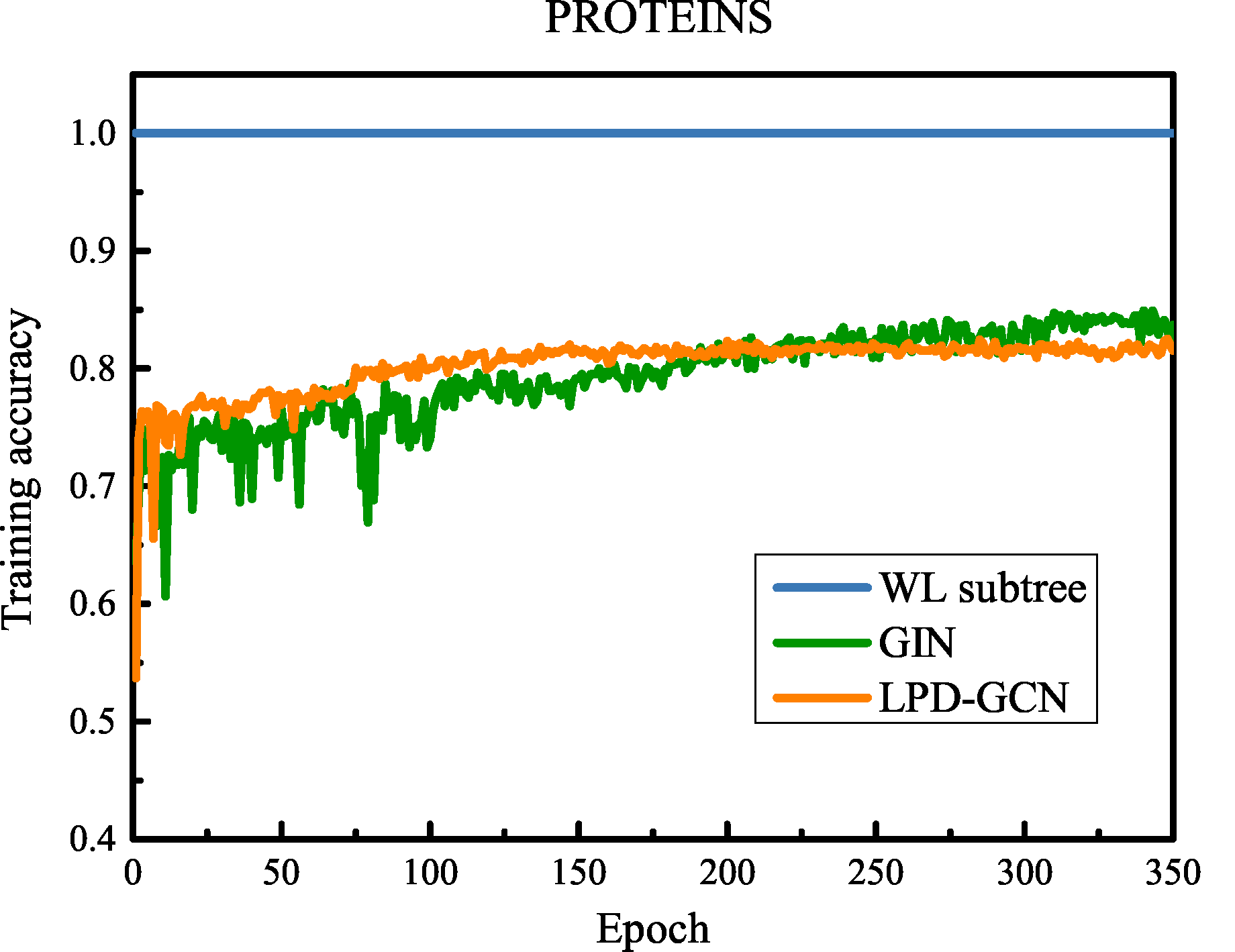}}
	\subfigure[NCI1]{
		\label{figure3:subfig:e}
		\includegraphics[scale=0.235]{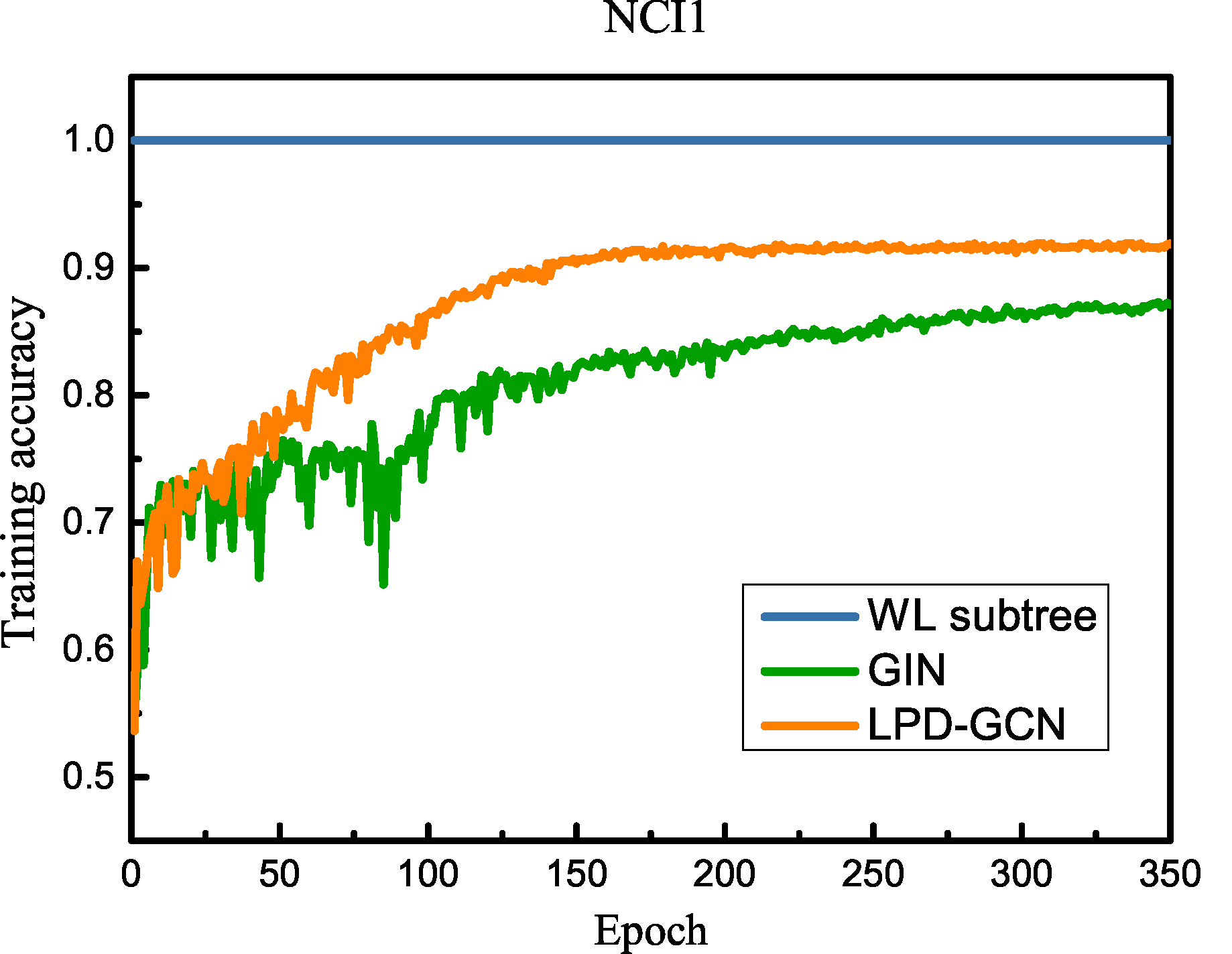}}
	\subfigure[D\&D]{
		\label{figure3:subfig:f}
		\includegraphics[scale=0.235]{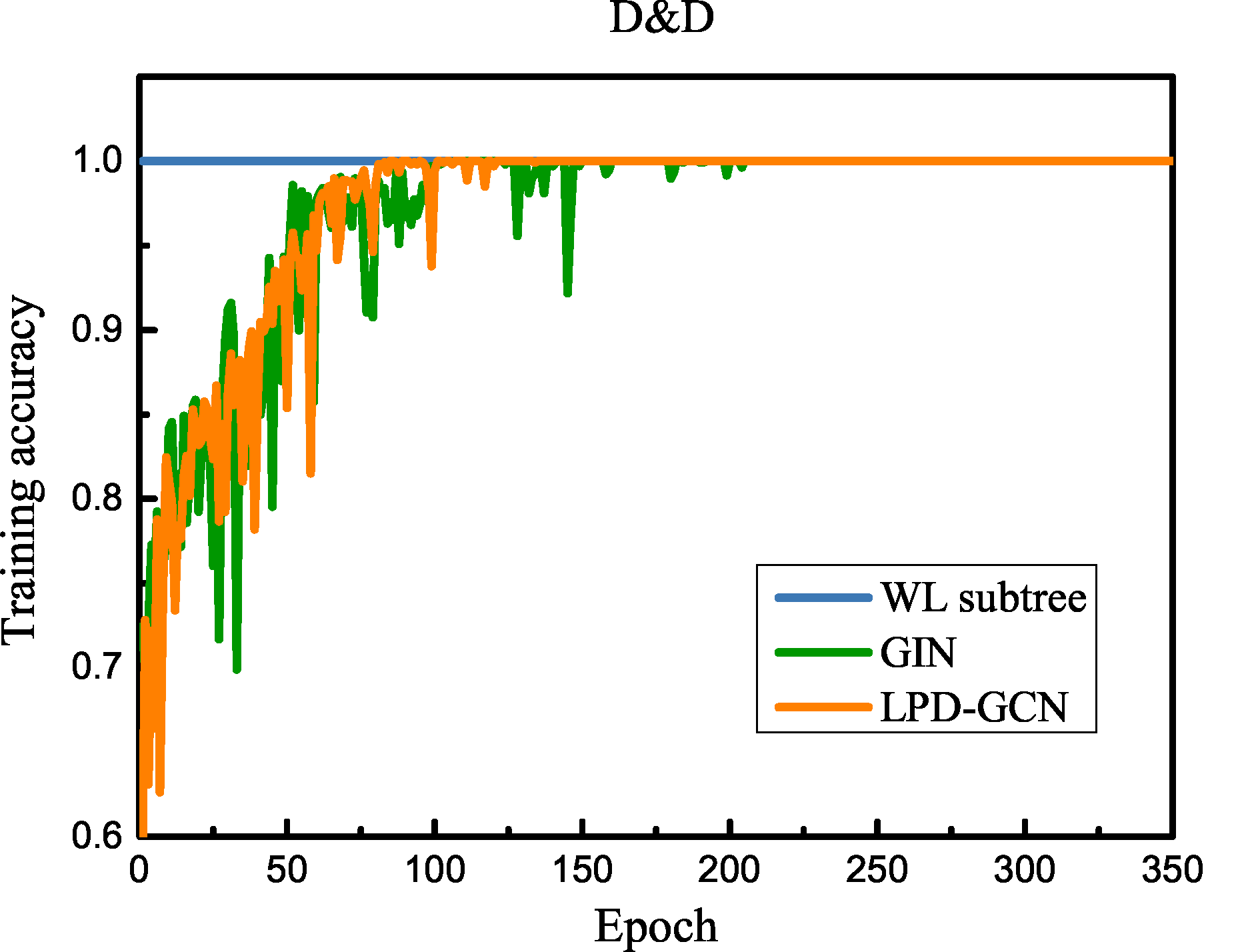}}
	\caption{Training set performance of the proposed LPD-GCN and two state-of-the-art baselines. Note that WL subtree curves always hold at 1.}
	\label{fig:figure3}
\end{figure}

\subsubsection{Training Set Performance}
The performance on training set reflects the expressive ability of models. In this section, we compare the experimental results on training sets to demonstrate the strong expressive power of our proposed LPD-GCN. Figure \ref{fig:figure3} shows training curves of LPD-GCN, as well as two state-of-the art baselines: WL subtree and GIN. From the results shown in Figure \ref{fig:figure3}, we have the following observations: In \cite{xu2018powerful}, the authors have demonstrated the strong discriminative power of the graph-kernel methods, which exceed those GCNs. This pattern aligns with our result that the WL test provides an upper bound for the representational capacity of the aggregation-based GCNs. For the MUTAG, ENZYMES, PTC and D\&D datasets, our proposed LPD-GCN is able to almost perfectly fit all the training sets, which achieves at most the same training accuracy as the WL subtree kernel. For PROTEINS and NCI1 datasets, LPD-GCN tends to fit the training sets better than GIN with faster convergence. Furthermore, for all the datasets, the training accuracy curves of LPD-GCN appear a smoother and steadier rising trend compared with GIN baseline, which indicates the training stage of LPD-GCN is more stable than GIN. To sum up, the above observations prove the powerful expressive capacity and stability of the proposed LPD-GCN.

\renewcommand{\arraystretch}{1.4}
\begin{table*}[t]\footnotesize
	\centering
	\caption{Test set classification accuracies (\%). We report the mean and std/dev of accuracies over a 10-fold stratified cross validation. The best results are highlighted with boldface. A method that significantly outperforms all methods is highlighted with boldface and asterisk. Gain denotes the accuracy improvement of our LPD-GCN compared with the baselines. }
	\label{tab:TestAcc}
	\begin{tabular}{llllllll}
		\hline
		\multirow{2}{*}{~} & \centering \multirow{2}{*}{METHOD} & \multicolumn{6}{c}{DATASET}   \tabularnewline
		\cline{3-8}
		~ & ~ & MUTAG & ENZYMES & PTC & PROTEINS & NCI1 & D\&D  \tabularnewline
		\hline

		\multirow{3}*{Kernel} & GRAPHLET & 81.7$\pm$2.1 & 41.0$\pm$1.0 & 54.7$\pm$1.4 & 71.7$\pm$0.6 & 62.3$\pm$0.3 & 74.9$\pm$1.1    \tabularnewline
		
		~ & SHORTEST-PATH & 81.7$\pm$2.5 & 42.3$\pm$1.5 & 58.9$\pm$2.4 & 76.4$\pm$0.5 &74.5$\pm$0.2 & 78.9$\pm$0.8 \tabularnewline
		
		~ & WL subtree & 90.4$\pm$5.7 & 61.25$\pm$3.4 & 59.9$\pm$4.3 & 75.0$\pm$3.1 & \textbf{86.0$\pm$1.8$^{*}$} & 78.3$\pm$0.6     \tabularnewline
		\hline
		\multirow{7}*{GCN} & Semi-GCN & 85.6$\pm$5.8 & $-$ & 64.2$\pm$4.3 & 76.0$\pm$3.2 & 80.2$\pm$2.0 & $-$  \tabularnewline
		
		~ & GraphSAGE & 85.1$\pm$7.5 & 54.25$\pm$6.0 & 63.9$\pm$7.7 & 75.9$\pm$3.2 & 77.7$\pm$1.5 & 75.4$\pm$3.0    \tabularnewline
		
		~ & PATCHY-SAN & 92.6$\pm$4.2 & $-$ & 60.0$\pm$4.8 & 75.9$\pm$2.8 & 78.6$\pm$1.9 & 76.3$\pm$2.4    \tabularnewline
		
		~ & SORTPOOL & 85.8$\pm$1.7 & 57.1$\pm$2.0 & 58.6$\pm$2.5 & 75.5$\pm$0.9 & 74.4$\pm$0.5 & 79.4$\pm$0.9      \tabularnewline			
		
		~ & GAT & 89.4$\pm$6.1 & $-$ & 66.7$\pm$5.1 & 74.7$\pm$2.2 & 75.2$\pm$3.3 & $-$      \tabularnewline
		
		~ & DIFFPOOL & $-$ & 62.5$\pm$5.6 & $-$ & 76.3$\pm$3.5 & $-$ & 80.6$\pm$3.5      \tabularnewline	
		
		~ & GIN & 89.0$\pm$6.0 & 53.3$\pm$4.7 & 63.7$\pm$8.2 & 75.9$\pm$3.8 & 82.7$\pm$1.6 & 80.9$\pm$2.7     \tabularnewline
		\hline	
		
		\multirow{3}*{Ours} & LPD-GCN(NoLFR) & \textbf{95.2$\pm$3.9$^{*}$} & 63.5$\pm$4.8 & 71.5$\pm$6.3 & 78.7$\pm$3.2 & 81.0$\pm$1.6 & 81.3$\pm$1.93    \tabularnewline
		
		~ & LPD-GCN(NoDC) & 94.5$\pm$5.8 & 65.3$\pm$7.2 & 73.2$\pm$6.2 & 79.0$\pm$3.0 & 81.8$\pm$5.5 & 82.6$\pm$4.0    \tabularnewline
		
		~ & LPD-GCN(NoGCA) & 94.2$\pm$4.3 & 61.7$\pm$5.1 & 71.7$\pm$4.5 & 78.3$\pm$2.2 & 78.7$\pm$1.6 & 82.1$\pm$2.5    \tabularnewline
		
		~ & LPD-GCN & 94.8$\pm$4.3 & \textbf{66.5$\pm$5.4$^{*}$} & \textbf{74.6$\pm$4.5$^{*}$} & \textbf{79.5$\pm$2.9$^{*}$} & 82.9$\pm$1.5 & \textbf{82.9$\pm$2.4}    \tabularnewline
		\hline
		~ & Gain & $\geq$~2.6 & $\geq$~4.0 & $\geq$~7.9 & $\geq$~3.2 & $-$ & $\geq$~2.0 \tabularnewline
		\hline
	\end{tabular}
\end{table*}

\subsubsection{Test Set Performance}
More importantly, to demonstrate the generalization ability of LPD-GCN, we further compare the test performance of LPD-GCN to these state-of-the-art graph classification baselines. The results shown in Table \ref{tab:TestAcc} provide positive answers to our motivating questions \textbf{Q1}, \textbf{Q2} and \textbf{Q3}. We can observe that our proposed LPD-GCN obtains the highest average performance and smallest std-dev among all GCN-based baselines, and we perform the wilcoxon rank sum test on results to show that our approach significantly outperform the other baselines on 4 out of 6 benchmark datasets, including MUTAG, ENZYMES, PTC and PROTEINS datasets. Especially for ENZYMES and PTC datasets, LPD-GCN relatively improves state-of-the-art baselines by at least 4.0\% and 7.9\% (accuracy), respectively. For other benchmark datasets like MUTAG, PROTEINS and D\&D, our LPD-GCN respectively yields at least 2.6\%, 3.2\% and 2.0\% (accuracy) gains over the 10 baselines. Although the WL subtree kernel achieves state-of-the-art performance in NCI1 dataset, our LPD-GCN outperforms most of the GCN and kernel baselines except for GIN, but our LPD-GCN can still achieve a comparable performance compared with GIN. The above observations demonstrate that the proposed LPD-GCN has better generalization ability than these GCN and kernel baselines. Interestingly, the proposed LPD-GCN achieves better results than it variants, i.e., LPD-GCN(NoLFR), LPD-GCN(NoDC) and LPD-GCN(NoGCA), which verify the proposed modifications on the model can help improve the performances. It is also observed that removing the dense connections in the model, LPD-GCN could be unstable to train, and there is a large variation in accuracy across different runs. Overall, LPD-GCN obtains a remarkable improvement on classification performance among the baseline models, which demonstrates it effectiveness and stability.

\subsection{Study on Local Feature Reconstruction}
\begin{figure}[htb!]
	\centering
	\includegraphics[scale=0.83]{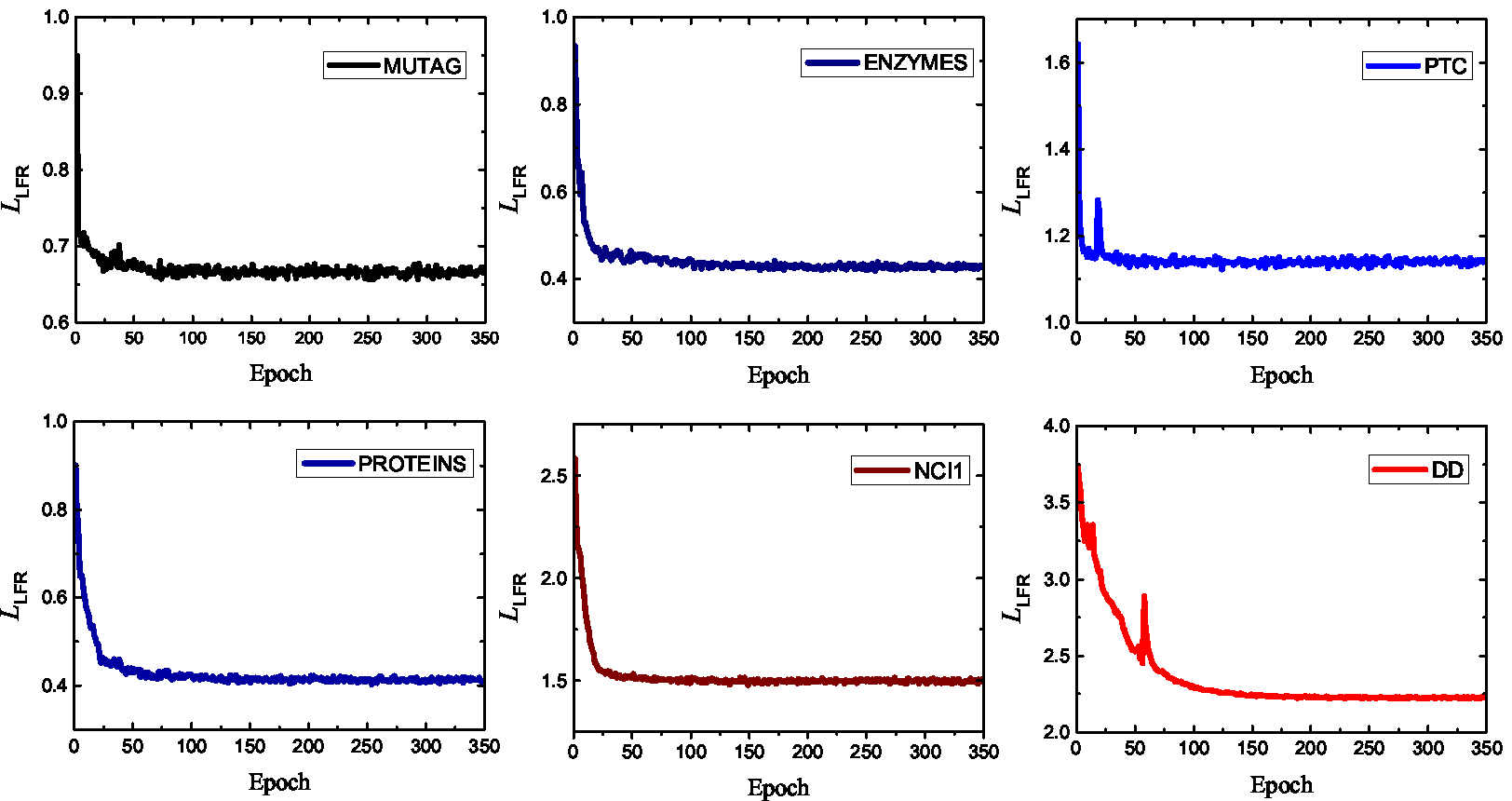}
	\caption{The decline curves of the local feature reconstruction loss during the training stage.}
	\label{fig:figure4}	
\end{figure}

To further support answering \textbf{Q2}, in Figure \ref{fig:figure4}, we present the true decline curves of the training loss of local feature reconstruction on all tested datasets. From the results shown in Figure \ref{fig:figure4}, we can observe that the local feature reconstruction loss sharply declines on all the datasets, which means that LPD-GCN can effectively reconstruct local feature informations and learns high-quality node representations. And then, the useful local information promotes the expressive ability of global representations. Except for D\&D dataset, the network scale of the other datasets is relatively small. Thus, the feature reconstruction loss achieves a quick convergence. For D\&D dataset, the average network scale reaches to 284, it converges slower than other datasets. The above observation certifies that the auxiliary supervision from local feature reconstruction loss can truly assist the goal of graph classification task. In next section, we will examine how to balance the contributions of local feature reconstruction loss and graph classification loss.

\subsection{Study on Hyper-parameter Sensitivity}
To address \textbf{Q4}, in this section, we analyze the hyper-parameter sensitivity of our LPD-GCN approach. Specifically, we evaluate the performance of LPD-GCN on graph classification tasks with respect to the trade-off parameter $\lambda$ and the dropout ratio. All hyper-parameters except the examined parameter are fixed for a fair comparison.

\begin{figure}[htb]
	\centering
	\subfigure[MUTAG]{
		\label{figure5:subfig:a}
		\includegraphics[scale=0.38]{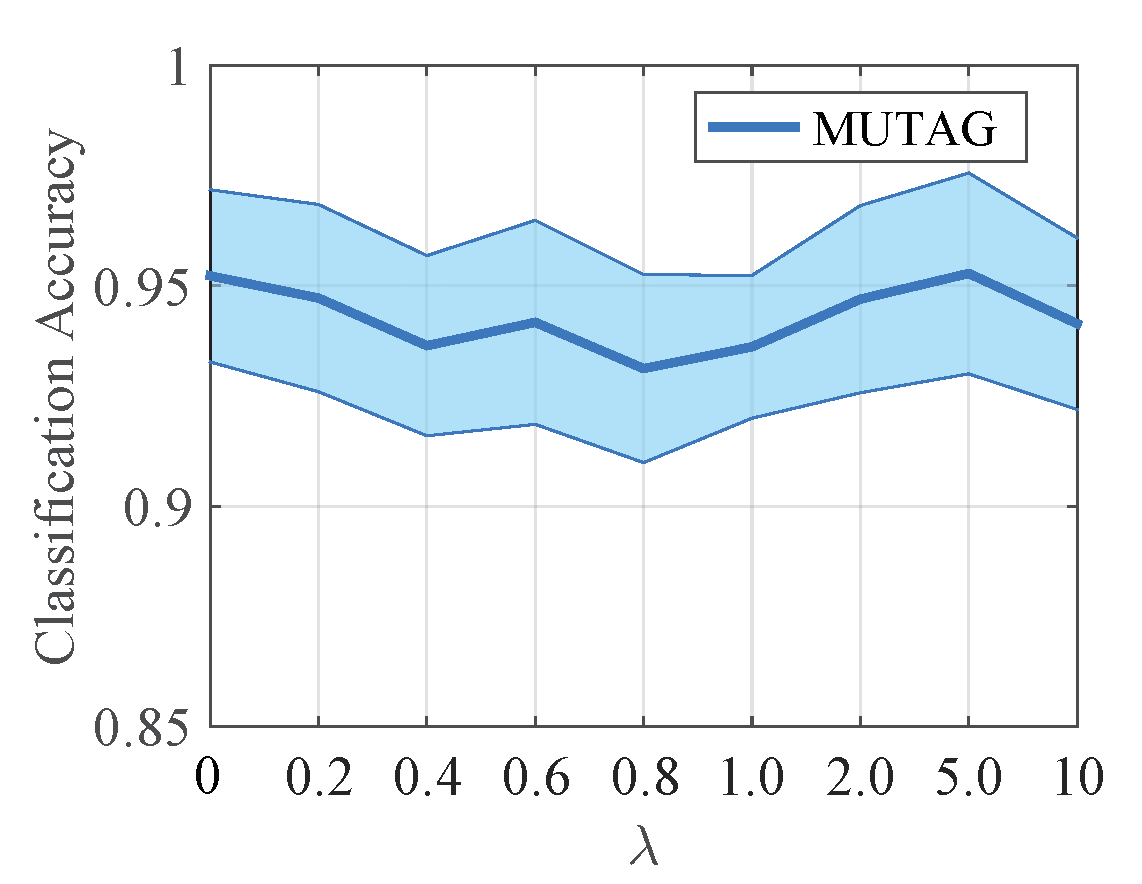}}
	\subfigure[ENZYMES]{
		\label{figure5:subfig:b}
		\includegraphics[scale=0.38]{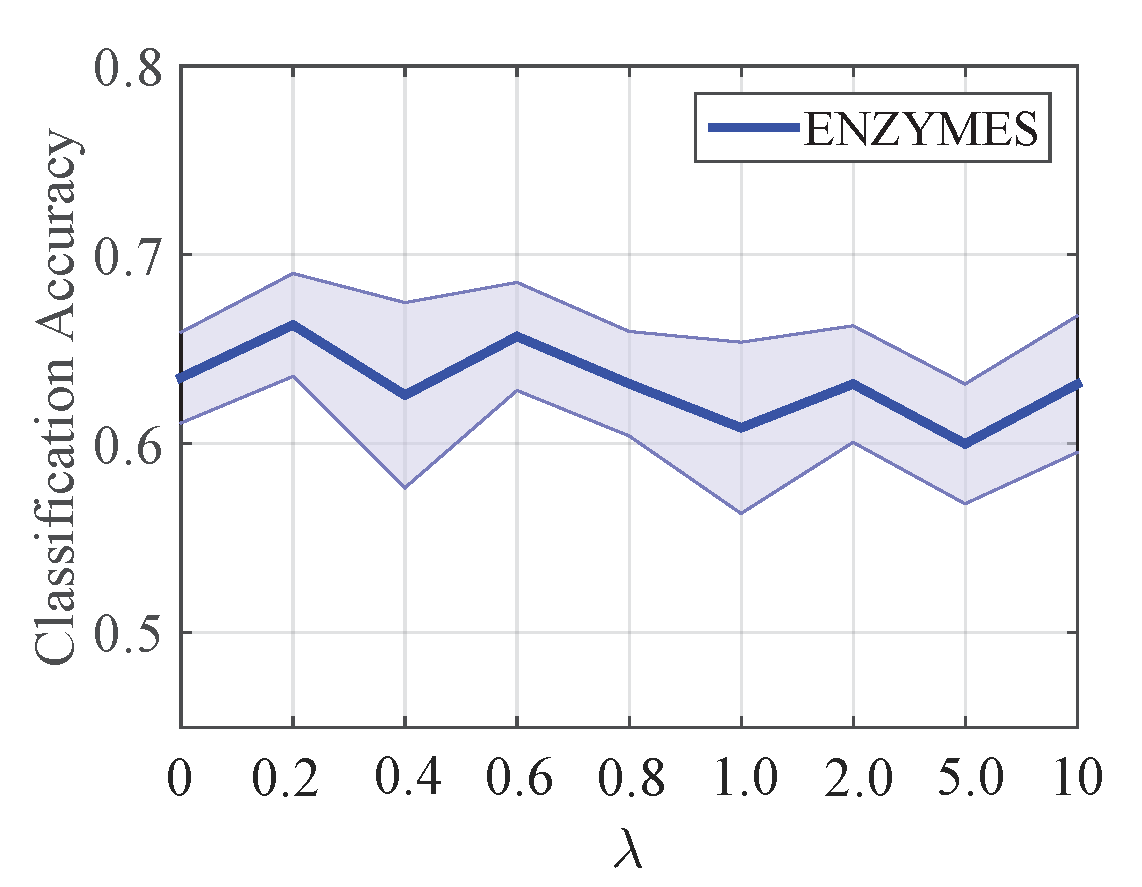}}
	\subfigure[PTC]{
		\label{figure5:subfig:c}
		\includegraphics[scale=0.38]{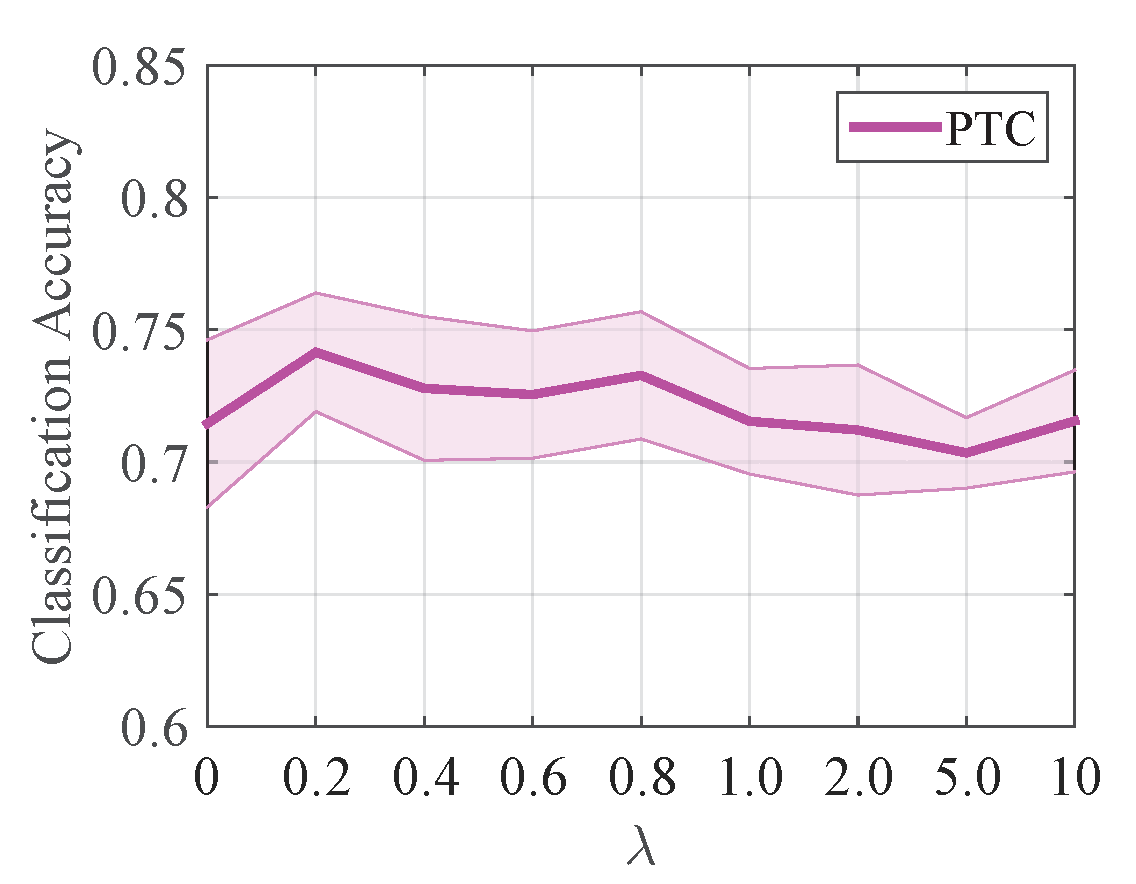}}
	\subfigure[PROTEINS]{
		\label{figure5:subfig:d}
		\includegraphics[scale=0.38]{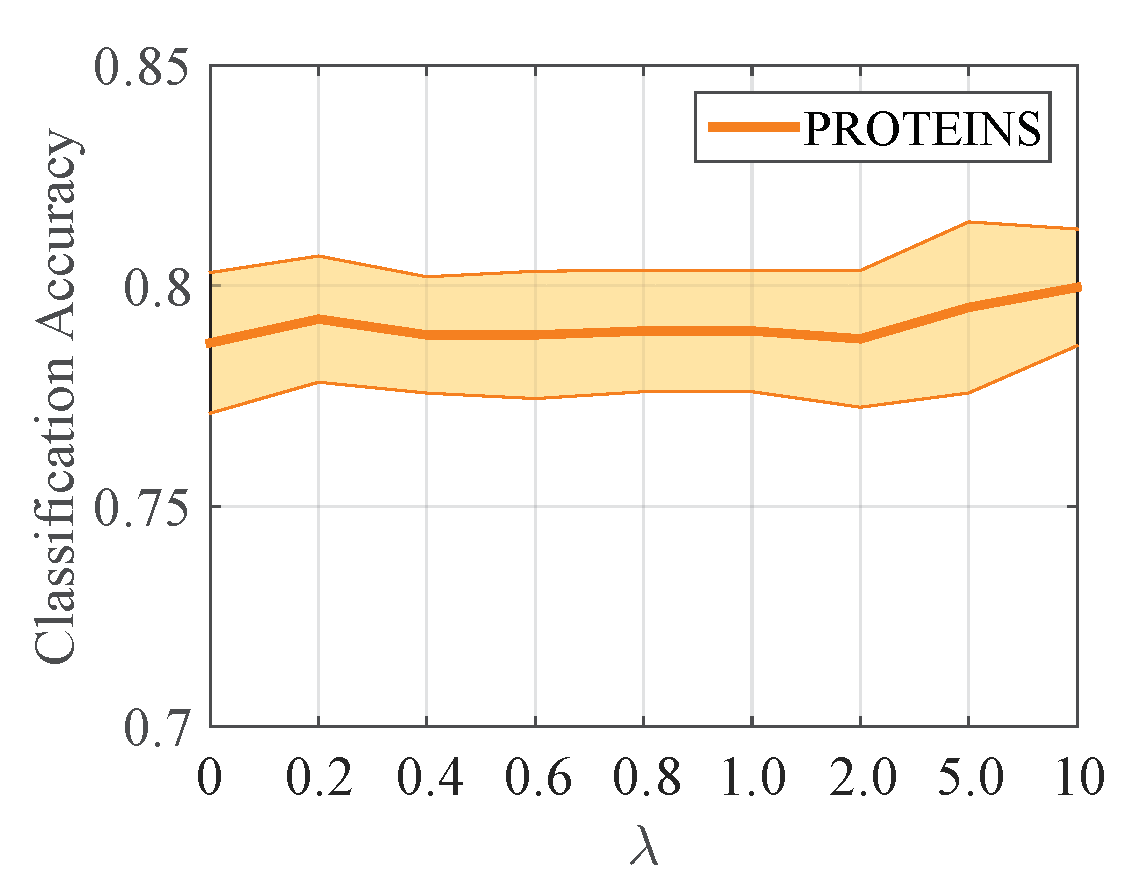}}
	\subfigure[NCI1]{
		\label{figure5:subfig:e}
		\includegraphics[scale=0.38]{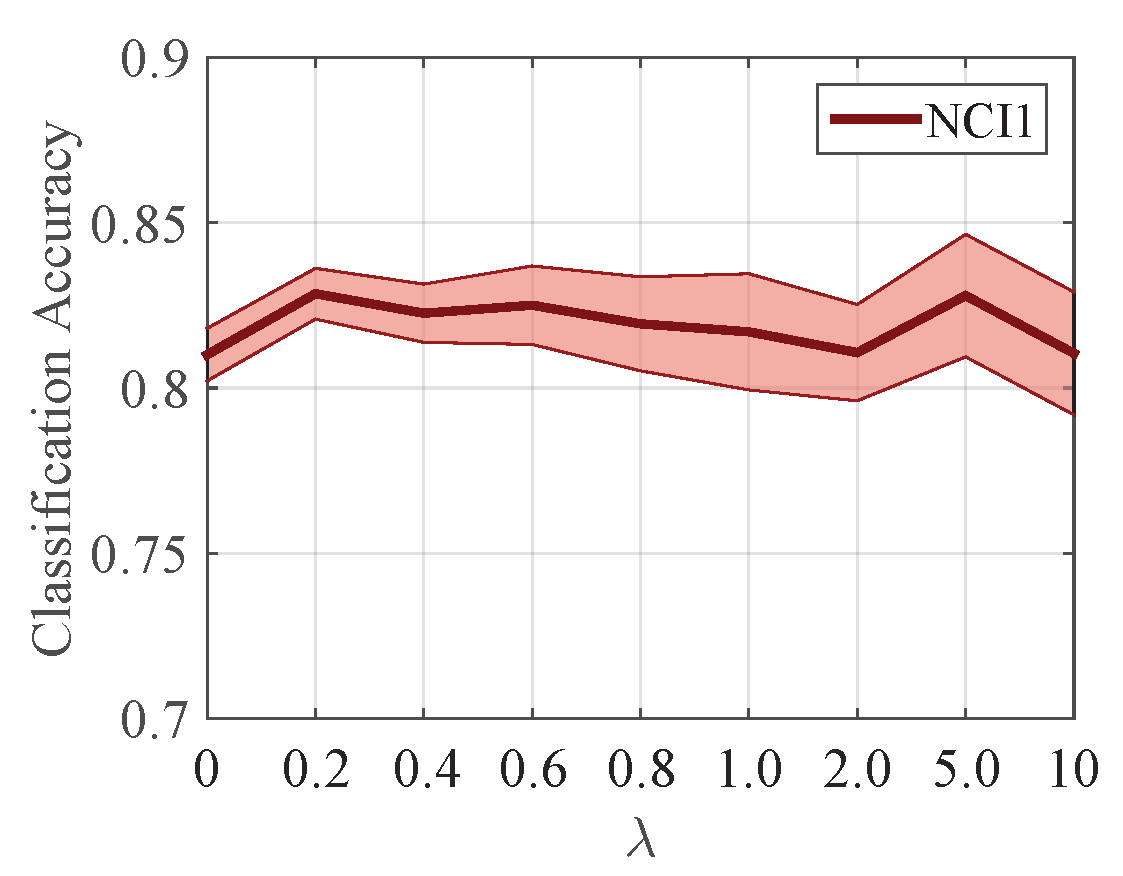}}
	\subfigure[D\&D]{
		\label{figure5:subfig:f}
		\includegraphics[scale=0.38]{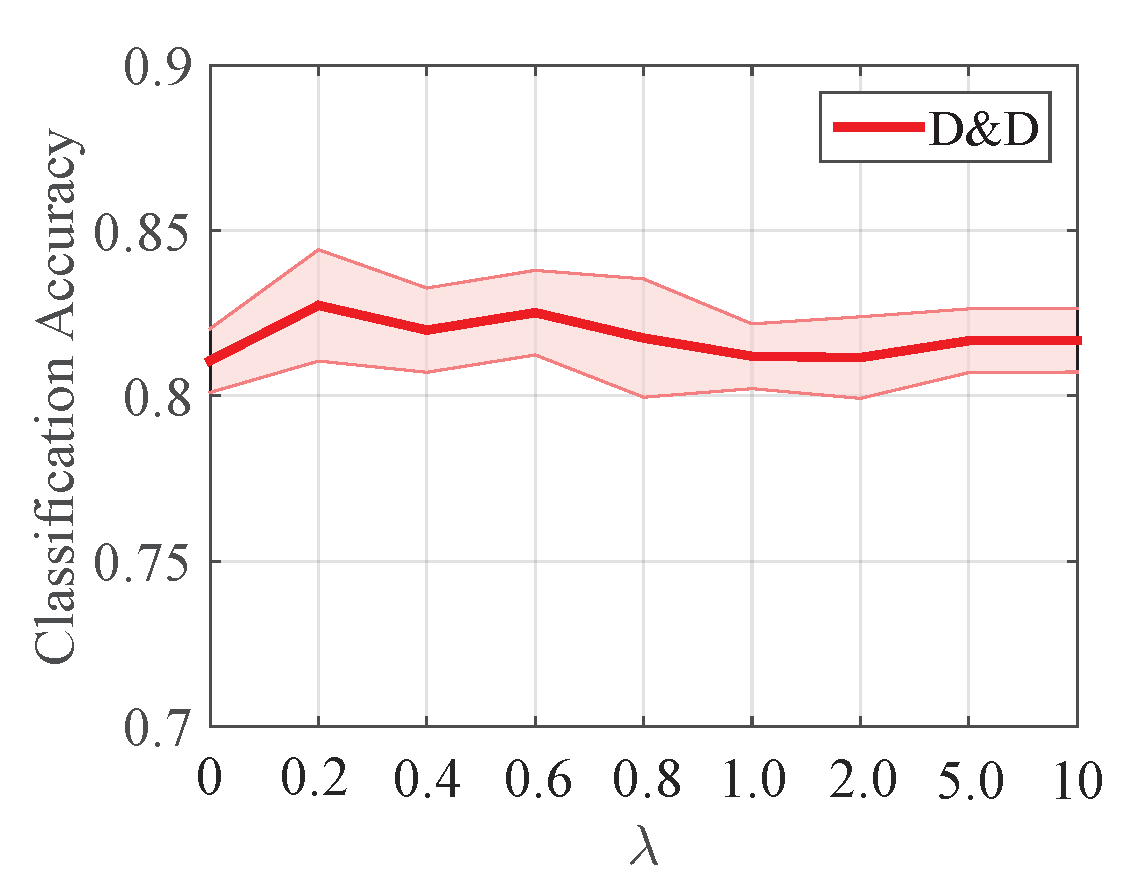}}
	\caption{Effects of the trade-off parameter $\lambda$ for LPD-GCN in terms of graph classification accuracy. We show the results on all the six test datasets. The shaded region denotes the half standard deviation on each side of the mean. It is worth noting that if $\lambda=0$ LPD-GCN turns to be the variant: LPD-GCN(NoLFR).}
	\label{fig:figure5}
\end{figure}

\textbf{Effect of the trade-off parameter $\lambda$.} We first discuss the influence of the trade-off parameter $\lambda$ in Equation (\ref{{TotalLoss}}). With different $\lambda$ values, the graph classification accuracy results on all the benchmark datasets are given in Figure \ref{fig:figure5}. First, it can be observed that adding the local feature reconstruction loss ($\lambda>0$) did improve the performance of graph classification task. Second, there is a trade-off between the local feature reconstruction loss and the graph classification loss, as too large $\lambda$ may incur noticeable deterioration in graph classification performance. Hence, $\lambda$ needs to be more carefully selected to balance the contributions of the local and global feature information. We test the experimental results of the trade-off parameter $\lambda \in \{0,0.2,0.4,0.6,0.8,1,2,5,10\}$. As shown in Figure \ref{fig:figure5}, we found that our model achieves the best performance on nearly all benchmarks when $\lambda$ is set to 0.2, except the MUTAG and PROTEINS datasets. Hence, we empirically set $\lambda$ to 0.2 in our experiments.

\begin{figure}[htb!]
	\centering
	\includegraphics[scale=0.36]{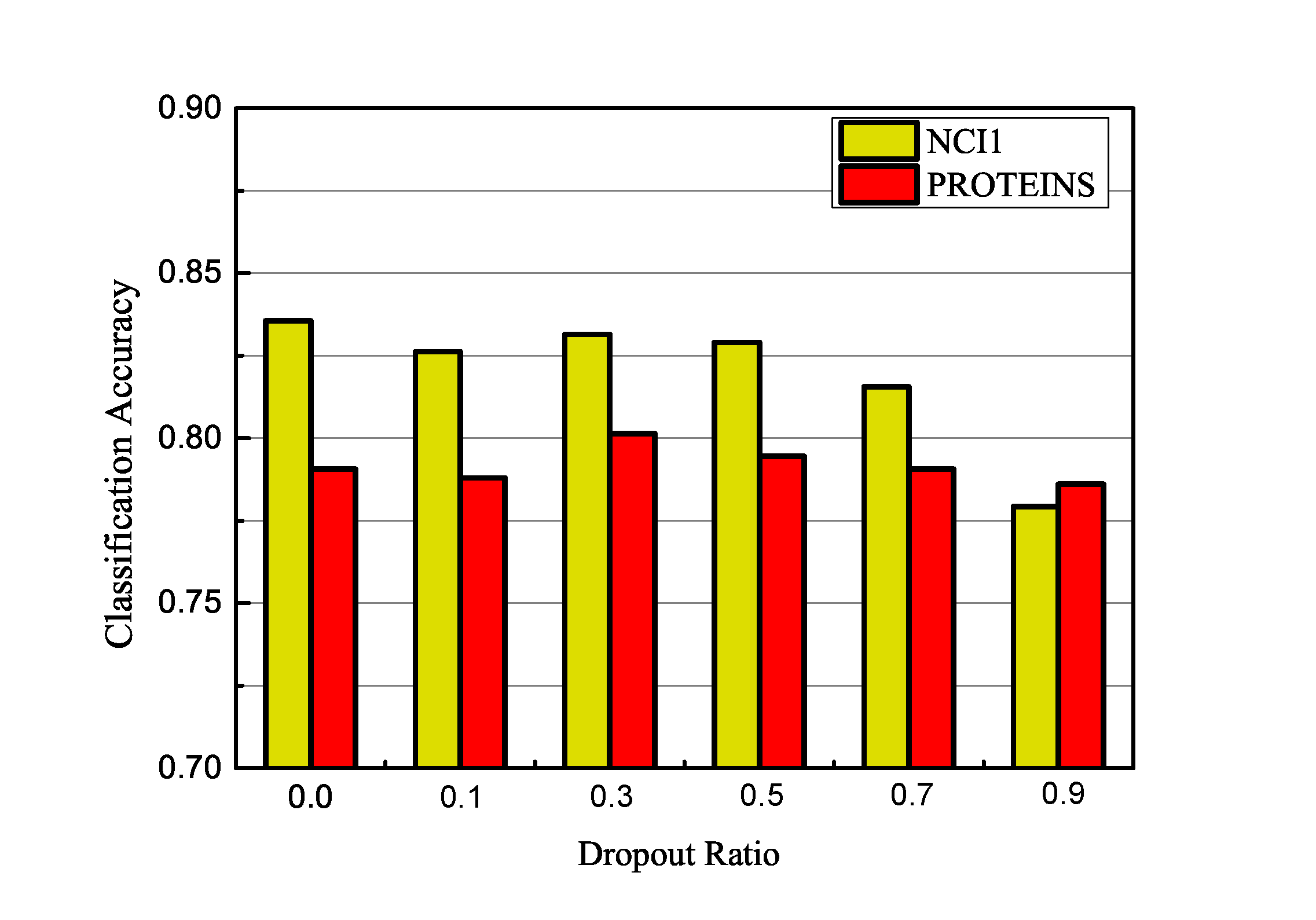}
	\caption{Effects of the dropout ratio for LPD-GCN in terms of graph classification accuracy. We show the results on two datasets: PROTEINS and NCI1. Here the trade-off parameter $\lambda$ is fixed to 0.2.}
	\label{fig:figure6}	
\end{figure}

\textbf{Effect of the dropout ratio.} Moreover, our LPD-GCN involves another important hyper-parameter: dropout ratio, which also dramatically affects the results of our model. In Figure \ref{fig:figure6}, we fix the trade-off parameter $\lambda$ to 0.2 and repeat the graph classification experiment to show how the different choices of dropout ratio $\in \{0,0.1,0.3,0.5,0.7,0.9\}$ would influence the performance of our model. Here we exhibit the results on the NCI1 and PROTEINS datasets as two examples. As shown in Figure \ref{fig:figure6}, it can be seen our LPD-GCN is more sensitive to different choice of dropout ratio on NCI1 dataset. However, the dropout ratio between 0 and 0.5 makes no obvious difference to the performance. The dropout ratio greater than 0.5 diminishes the accuracy sharply. On PROTEINS dataset, the model achieves the optimal performance when the dropout ratio is in the range from 0.3 to 0.5. Based on these observations, we find that the dropout ratio of 0.5 is a proper choice in terms of the overall performance on the tested datasets.

\subsection{Study on Scalability}
\begin{figure}[htb]
	\centering
	\subfigure[MUTAG]{
		\label{figure7:subfig:a}
		\includegraphics[scale=0.24]{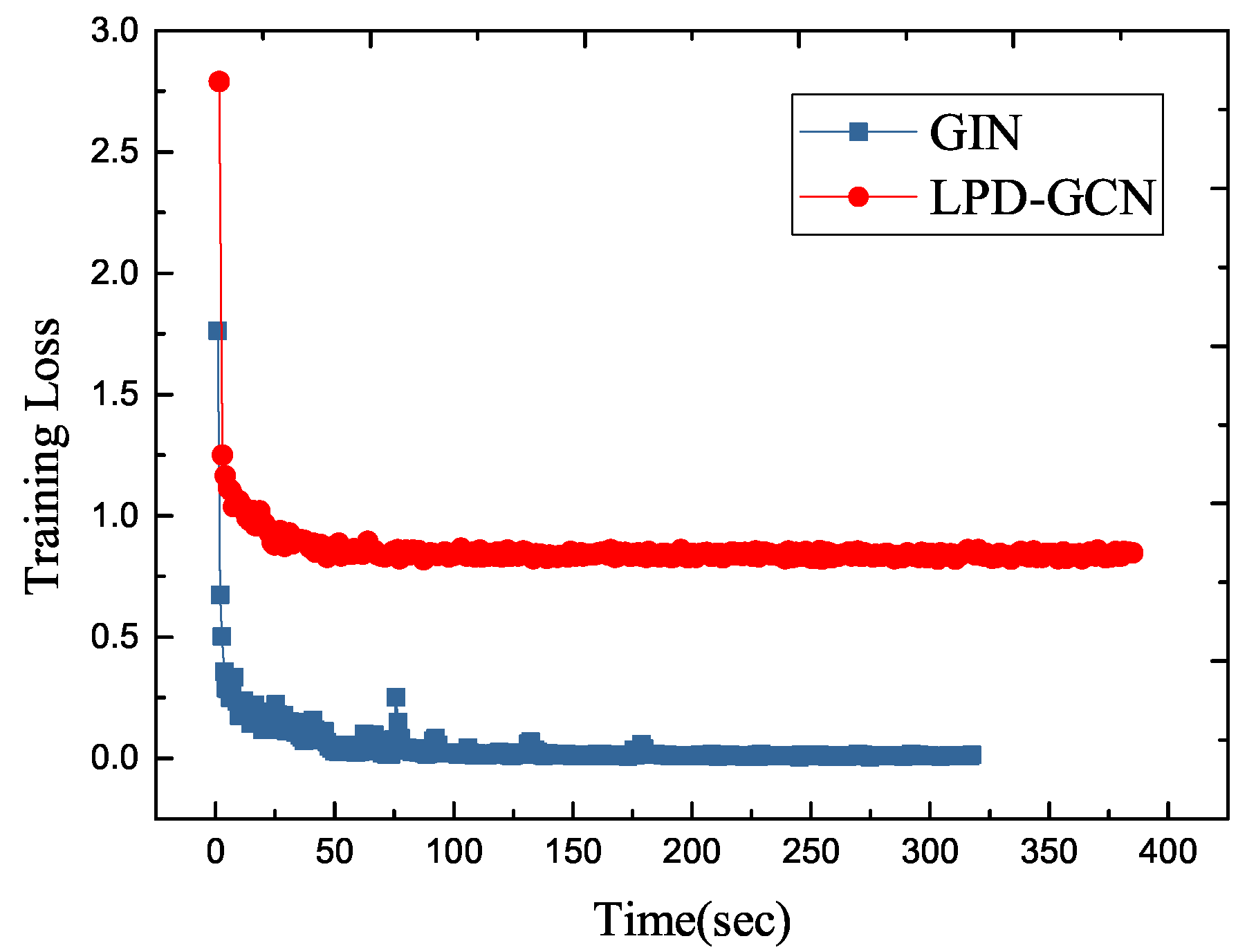}}
	\subfigure[ENZYMES]{
		\label{figure7:subfig:b}
		\includegraphics[scale=0.24]{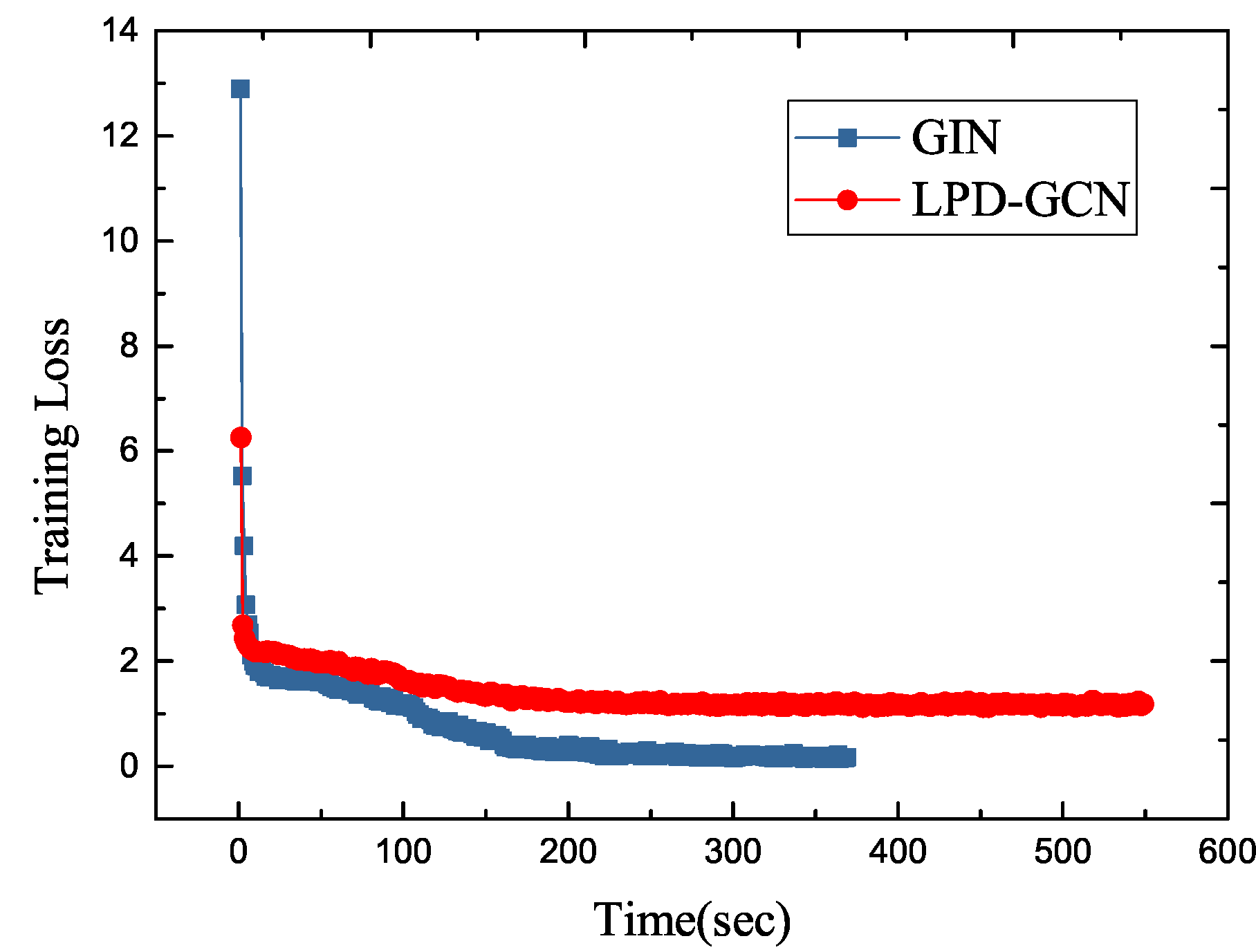}}
	\subfigure[PTC]{
		\label{figure7:subfig:c}
		\includegraphics[scale=0.24]{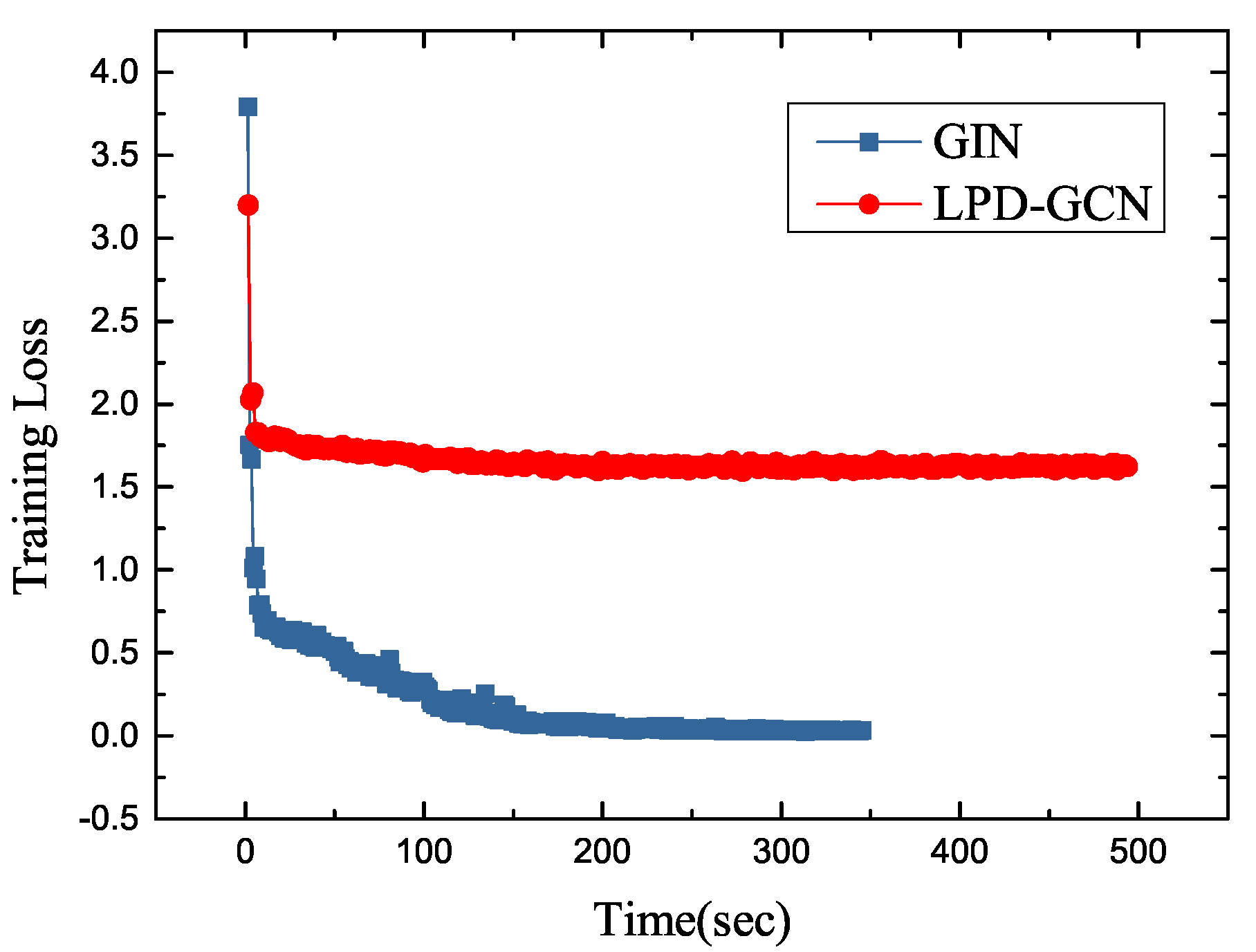}}
	\subfigure[PROTEINS]{
		\label{figure7:subfig:d}
		\includegraphics[scale=0.24]{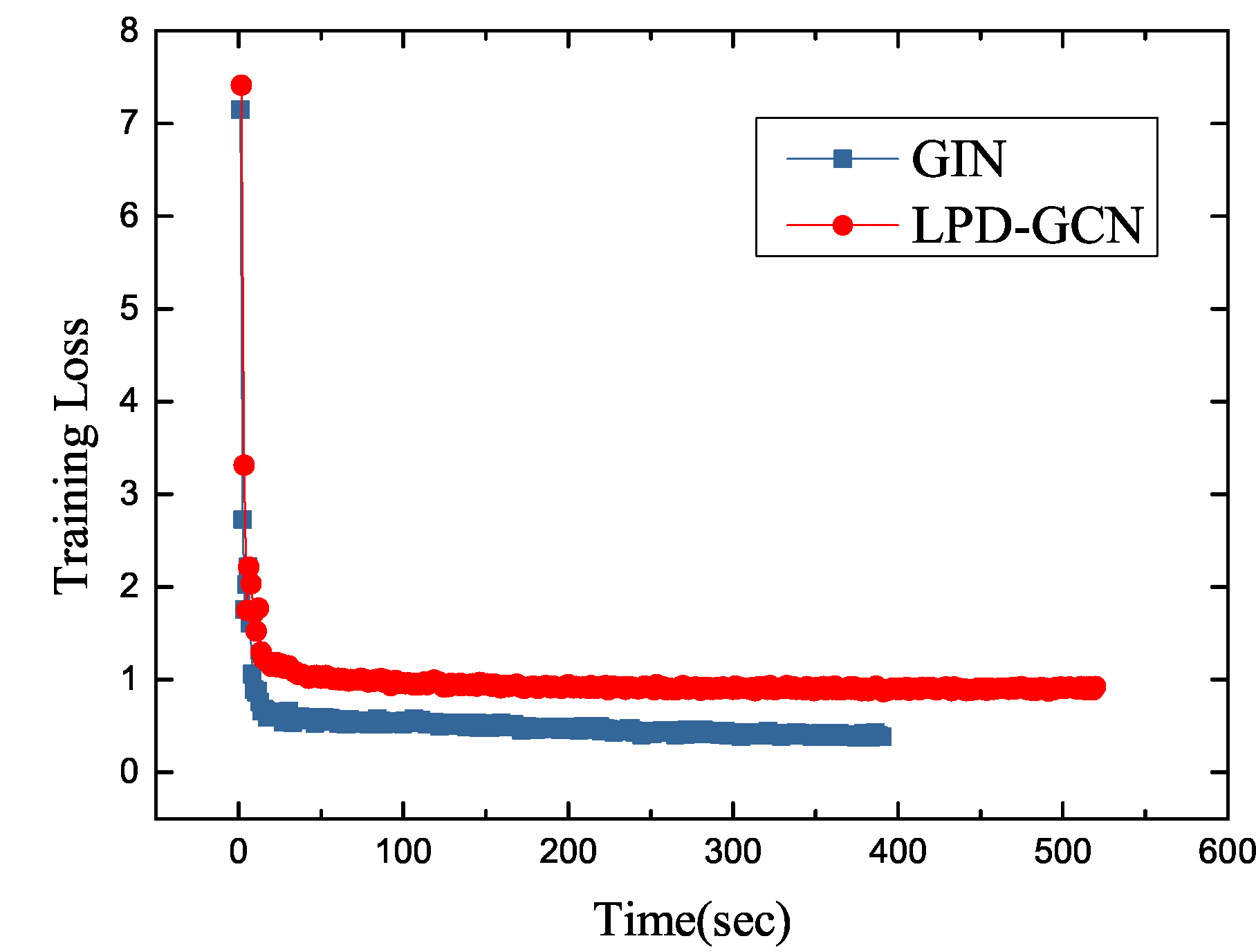}}
	\subfigure[NCI1]{
		\label{figure7:subfig:e}
		\includegraphics[scale=0.24]{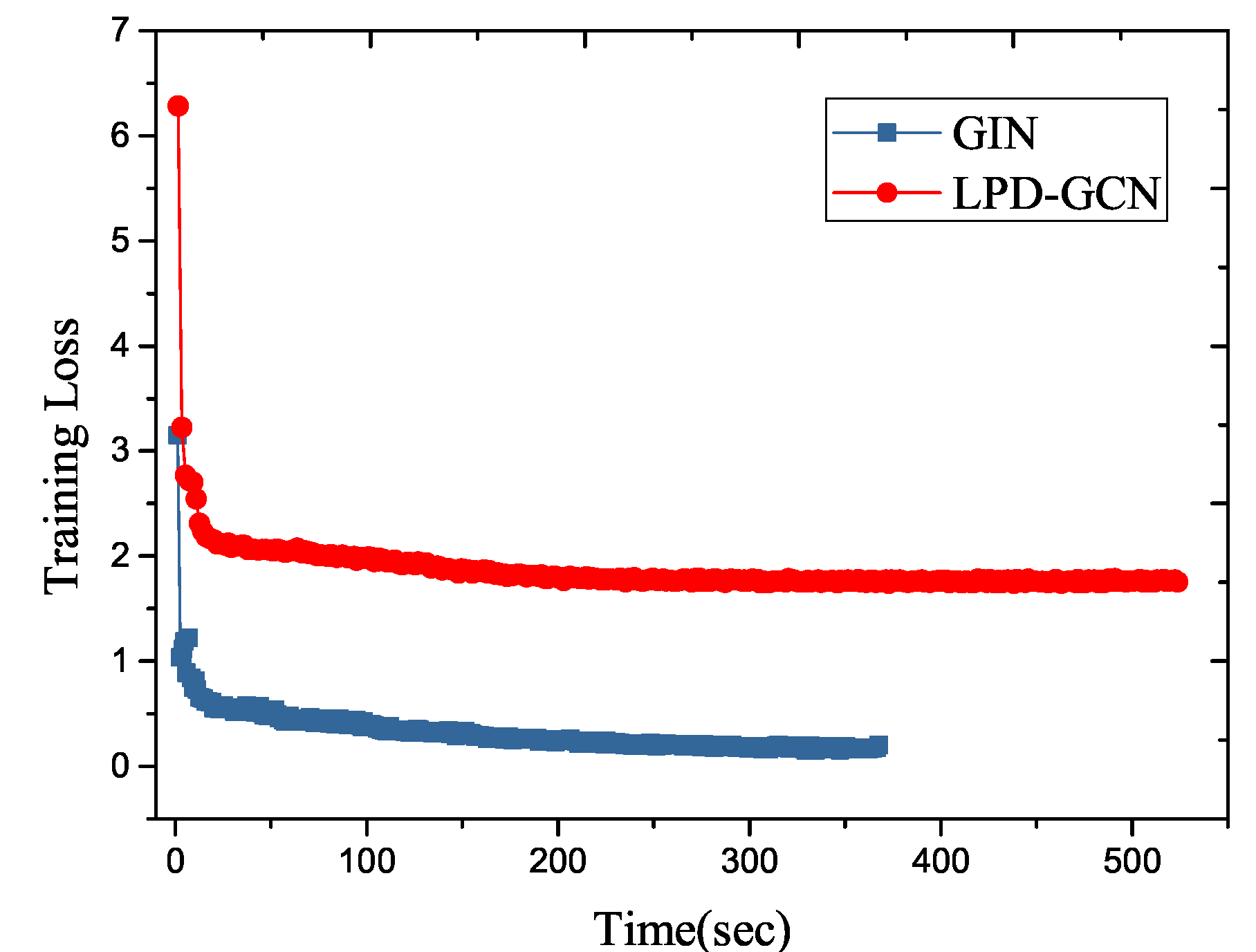}}
	\subfigure[D\&D]{
		\label{figure7:subfig:f}
		\includegraphics[scale=0.24]{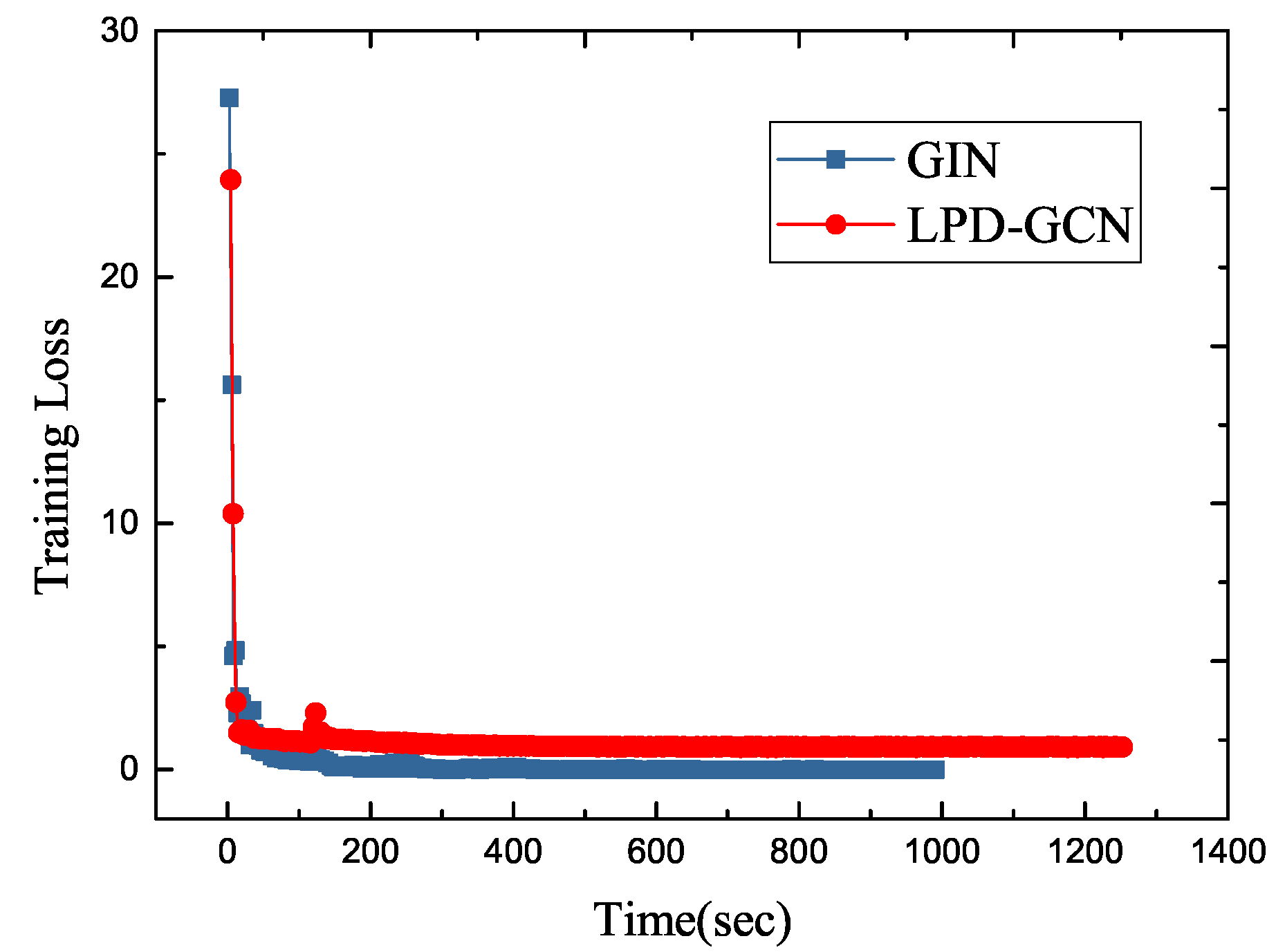}}
	\caption{Time usages of the proposed LPD-GCN and GIN.}
	\label{fig:figure7}
\end{figure}

In this section, we evaluate the scalability of our proposed LPD-GCN. In Figure \ref{fig:figure7}, we compare the training time usages between LPD-GCN and GIN baseline on all the benchmark datasets. Specifically, we run a total of 350 epochs and record the loss decline curves according to their training time. As we can see from Figure \ref{fig:figure7}, compared with GIN baseline, our LPD-GCN occupies relatively more computational costs for each training epoch, because of the additional decoder for the local feature reconstruction and the dense connections between convolutional layers. However, LPD-GCN takes less training time to reach convergences on nearly all the benchmark datasets. This shows that through adding the local feature-reconstruction regularization, our model can achieve faster convergence compared with GIN baseline.

\section{Conclusion and Future Work}
In this paper, we develop a locality preserving dense graph convolutional network architecture with graph context-aware node representations for graph classification. Based upon the general graph convolutional network framework, our model constructs an auxiliary local node-feature reconstruction loss to assist the goal of graph classification task. Besides, our model explores a dense connectivity pattern to flexibly leverage information from differing locality and improve model generalization. Moreover, we design a simple architecture to help local-global information flow across the model and introduce a self-attention mechanism for generating the final graph-level representation. Experimental results demonstrate that the proposed model achieves significant improvement of classification accuracies on 4 out of 6 benchmark datasets compared to state-of-the-art baseline models.

In our future research, we will extend our model to handle non-attributed graph classification problem. Additionally, we are interested in exploring a generic model which not only performs on graph classification but also is applied to other applications, e.g., recommender system, and one-shot or zero-shot learning and point-cloud classification, etc. We expect our model to provide new insights for training deeper GCNs.

\bibliographystyle{elsarticle-harv}
\bibliography{MyReferenceLibABR}

\end{document}